\def\paperlanguage{} 

\documentclass{ieeeaccess}

\usepackage{tabularx}

\usepackage{hyperref}
\usepackage{breakurl}
\hypersetup{
    colorlinks=true,
    linkcolor=black,
    citecolor=black,
    urlcolor=black,
    bookmarks=true,
    bookmarksnumbered=true
}

\newcommand{\secref}[1]{Section~\ref{#1}}
\newcommand{\tabref}[1]{{Table~\ref{#1}}}
\newcommand{\figref}[1]{{Fig.~\ref{#1}}}

\newcommand{\switchlanguage}[2]{%
  \ifx\paperlanguage\empty%
  #1%
  \else%
  #2%
  \fi%
}

\usepackage{amsthm}

\newcommand{\ctext}[1]{\raise0.2ex\hbox{\textcircled{\scriptsize{#1}}}}

\graphicspath{{figs/}{./figs/}{figures/}{./figures/}}

\usepackage{cite}
\usepackage{amsmath,amssymb,amsfonts}
\usepackage{algorithmic}
\usepackage{graphicx}
\usepackage{textcomp}

\usepackage{bm}
\makeatletter
\AtBeginDocument{\DeclareMathVersion{bold}
\SetSymbolFont{operators}{bold}{T1}{times}{b}{n}
\SetSymbolFont{NewLetters}{bold}{T1}{times}{b}{it}
\SetMathAlphabet{\mathrm}{bold}{T1}{times}{b}{n}
\SetMathAlphabet{\mathit}{bold}{T1}{times}{b}{it}
\SetMathAlphabet{\mathbf}{bold}{T1}{times}{b}{n}
\SetMathAlphabet{\mathtt}{bold}{OT1}{pcr}{b}{n}
\SetSymbolFont{symbols}{bold}{OMS}{cmsy}{b}{n}
\renewcommand\boldmath{\@nomath\boldmath\mathversion{bold}}}
\makeatother

\def\BibTeX{{\rm B\kern-.05em{\sc i\kern-.025em b}\kern-.08em
    T\kern-.1667em\lower.7ex\hbox{E}\kern-.125emX}}

\begin{document}
\history{Date of publication xxxx 00, 0000, date of current version xxxx 00, 0000.}
\doi{10.1109/ACCESS.2024.0429000}
\bstctlcite{BSTcontrol}

\title{Efficient Robot Design with Multi-Objective Black-Box Optimization and Large Language Models}

\author{
  Kento Kawaharazuka\authorrefmark{1},
  Yoshiki Obinata\authorrefmark{1},
  Naoaki Kanazawa\authorrefmark{2},
  Haoyu Jia\authorrefmark{1},
  and Kei Okada\authorrefmark{1}
}

\address[1]{Department of Mechano-Informatics, The University of Tokyo, Bunkyo-ku, Tokyo 113-8656 JAPAN ([kawaharazuka, obinata, kanazawa, jia, k-okada]@jsk.imi.i.u-tokyo.ac.jp)}
\tfootnote{This research was partially supported by JST FOREST under Grant Number JPMJFR232N.}

\markboth
{K. Kawaharazuka \headeretal: Efficient Robot Design Optimization}
{K. Kawaharazuka \headeretal: Efficient Robot Design Optimization}

\corresp{Corresponding author: Kento Kawaharazuka (e-mail: kawaharazuka@jsk.imi.i.u-tokyo.ac.jp).}

\begin{abstract}
  \switchlanguage%
  {%
    Various methods for robot design optimization have been developed so far.
    These methods are diverse, ranging from numerical optimization to black-box optimization.
    While numerical optimization is fast, it is not suitable for cases involving complex structures or discrete values, leading to frequent use of black-box optimization instead.
    However, black-box optimization suffers from low sampling efficiency and takes considerable sampling iterations to obtain good solutions.
    In this study, we propose a method to enhance the efficiency of robot body design based on black-box optimization by utilizing large language models (LLMs).
    In parallel with the sampling process based on black-box optimization, sampling is performed using LLMs, which are provided with problem settings and extensive feedback.
    We demonstrate that this method enables more efficient exploration of design solutions and discuss its characteristics and limitations.
  }%
  {%
    これまで様々なロボット設計最適化の手法が開発されてきた.
    それらは数値的な最適化からブラックボックス最適化まで多種多様である.
    数値的な最適化は高速な一方で複雑な構造や離散的な値が含まれるケースには向かず, ブラックボックス最適化が頻繁に利用されている.
    しかし, ブラックボックス最適化はサンプリング効率が低く良い解を得るまでに時間がかかる.
    そこで本研究では, このブラックボックス最適化に基づくロボット身体設計を, 大規模言語モデル(LLM)によって, より効率化する手法を提案する.
    ブラックボックス最適化に基づくサンプリングと並行して, 問題設定や多数のフィードバックを与えたLLMによるサンプリングを実行する.
    本手法がより効率的に設計解を探索できることを示すと同時に, その特性や限界などについて議論する.
  }%
\end{abstract}


\titlepgskip=-21pt

\maketitle

\section{INTRODUCTION}\label{sec:introduction}
\switchlanguage%
{%
  The diversity of robot designs has been increasing, and their body structures, particularly joint configurations, differ significantly between robots.
  There are also an increasing number of modular robots that can change their bodies according to tasks or user preferences \cite{alattas2019modular, romiti2022reconfigurable}.

  As various robot configurations are being constructed, many studies have been developed to identify favorable configurations and optimize the design of body structures.
  For example, in \cite{yang2000modular}, the number and types of modules, as well as their relative positions, have been optimized using a genetic algorithm to satisfy the target operation points for industrial robots.
  In \cite{xiao2016designopt}, a multi-objective optimization of motor specifications and gear ratios for a general 6-DOF manipulator has been performed using a genetic algorithm to minimize body weight and maximize manipulability.
  In \cite{zhao2020robogrammar}, the joint configuration and link lengths of a modular robot capable of traversing uneven terrain have been optimized using a graph heuristic search.
  In \cite{liu2020modular} an exhaustive search has been configured to design a modular robot that can execute tasks while satisfying joint angle limits, torque constraints, and collision avoidance.
  In \cite{hu2022modular}, the design of a modular robot has been optimized by generating diverse body structures using a Generative Adversarial Network (GAN).
  In \cite{lei2024optimization}, evolutionary strategies have been employed to optimize the module configuration to satisfy the commanded trajectory while minimizing manipulability and required torque.
  The work in \cite{kawaharazuka2023autodesign} has presented Pareto-optimal solutions that minimize position error and required torque through multi-objective optimization using a genetic algorithm for body adaptation to tasks and user preferences.
  In \cite{kawaharazuka2024slideopt}, body design optimization has been extended to include not only rotary joints but also linear joints.

  As illustrated, various methods such as genetic algorithms, evolutionary strategies, exhaustive search, heuristic search, and GANs have been utilized for optimizing the body design of diverse robots.
  Among these, research using black-box optimization (BBO), particularly genetic algorithms, is prevalent due to its applicability to complex problem settings, including optimizations involving both continuous and discrete variables as well as multi-objective optimization \cite{yang2000modular, xiao2016designopt, kawaharazuka2023autodesign, kawaharazuka2024slideopt, vanteddu2024cadurdf}.
  Although gradient-free black-box optimization is applicable to any problem, it has the drawback of requiring a substantial number of samples and time to generate high-quality solutions.
}%
{%
  現在はロボットの形が多様化しており, その身体構造, 特に関節配置等はロボットごとに大きく異なる.
  モジュラーロボットのように, タスクやユーザの好みに応じて身体を変化させることのできる構成も増えてきた\cite{alattas2019modular, romiti2022reconfigurable}.

  様々な構成のロボットが構築されるに従って, どのような構成が良いのか, 身体構造の設計を最適化する研究が多く開発されてきた.
  \cite{yang2000modular}では, 産業用ロボットに向け, 所望の動作点を満たすモジュールの数と種類, モジュール間の相対的な位置を遺伝的アルゴリズムにより最適化した.
  \cite{xiao2016designopt}では, 身体重量の最小化とmanipulabilityの最大化に基づき, 一般的な6自由度マニピュレータのモータとギア比を遺伝的アルゴリズムにより多目的最適化した.
  \cite{zhao2020robogrammar}では, 不整地を走行可能なモジュールロボットの関節配置やリンク長に関する最適化をGraph Heuristic Searchにより行った.
  \cite{liu2020modular}では, 総当りにより関節角度制限やトルク制限, 衝突回避等を満たしつつタスクを実行可能なモジュールロボットの設計最適化を行った.
  \cite{hu2022modular}では, Generative Adversarial Network (GAN)を用いた多様な身体生成と不整地走行可能なモジュールロボットの設計最適化を行った.
  \cite{lei2024optimization}では, 指令軌道を満たしつつmanipulabilityと必要トルクを最小化するモジュール配置を進化戦略により最適化した.
  \cite{kawaharazuka2023autodesign}では, タスクやユーザの好みに応じて身体を変化させるために, 遺伝的アルゴリズムを用いた多目的最適化により位置誤差と必要トルクを最小化するパレート解を提示した.
  \cite{kawaharazuka2024slideopt}は一般的な回転関節だけでなく, 直動関節まで含んだ身体設計の最適化を行っている.
  これらのように, 遺伝的アルゴリズム, 進化戦略, 総当り, ヒューリスティック探索, GANまで, 様々な手法を用いて多様なロボットの身体設計最適化が行われてきている.
  なかでも, 連続値と離散値を含む最適化や多目的な最適化など, 複雑な問題設定にも適用可能なブラックボックス最適化, 特に遺伝的アルゴリズムを用いた研究が盛んである\cite{yang2000modular, xiao2016designopt, kawaharazuka2023autodesign, kawaharazuka2024slideopt, vanteddu2024cadurdf}.
  勾配を用いないブラックボックス最適化は, いかなる問題にも適用可能な一方で, 良い解を生成するまでにはかなりのサンプリング数と時間がかかるという問題がある.
}%

\begin{figure}[t]
  \centering
  \includegraphics[width=0.98\columnwidth]{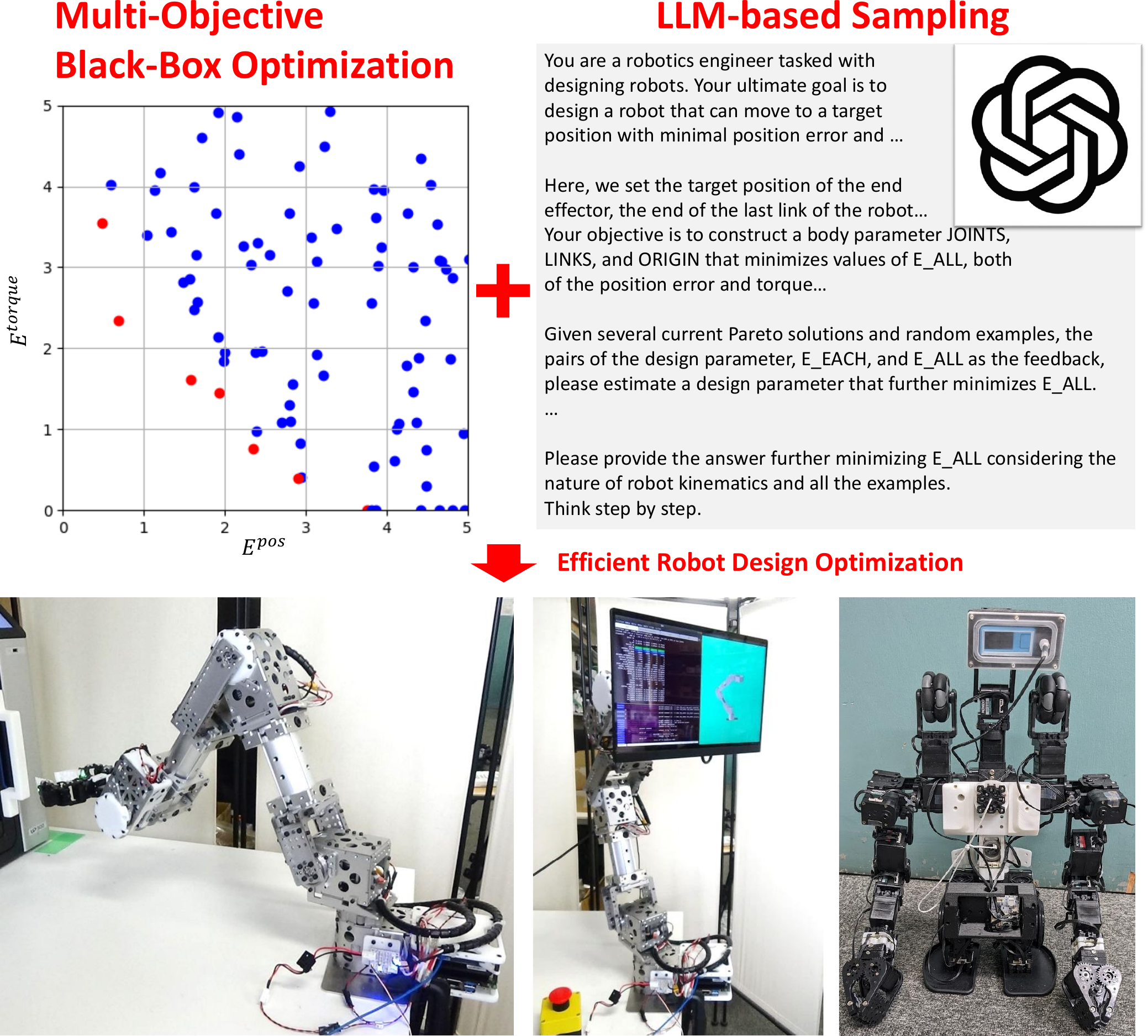}
  \vspace{-1.0ex}
  \caption{The concept of this study: high efficiency robot design with multi-objective black-box optimization and LLM-based sampling.}
  \vspace{-1.0ex}
  \label{figure:concept}
\end{figure}

\begin{figure*}[t]
  \centering
  \includegraphics[width=2.0\columnwidth]{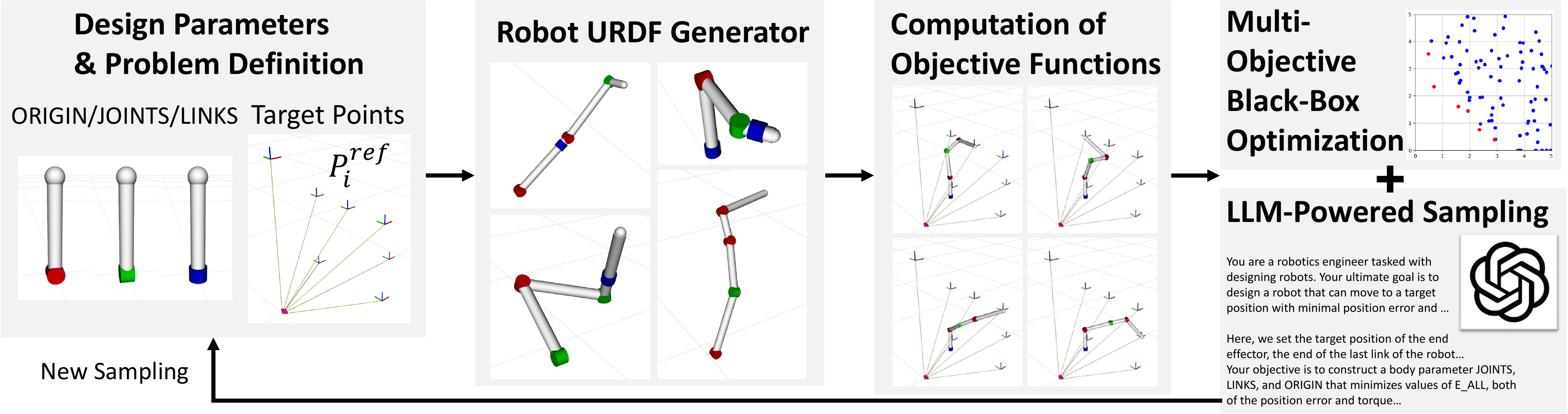}
  \vspace{-1.0ex}
  \caption{The overview of the proposed robot design optimization. We set the robot design parameters, automatically generate the robot URDF, compute the objective functions, and perform multi-objective black-box optimization and LLM-based sampling.}
  \vspace{-1.0ex}
  \label{figure:overview}
\end{figure*}

\switchlanguage%
{%
  As an alternative approach, examples of using large language models (LLMs) \cite{brown2020gpt3, achiam2023gpt4, kawaharazuka2024foundation} for body design have begun to emerge.
  In \cite{stella2023design}, it is qualitatively demonstrated that LLMs can effectively assist in robot design.
  In \cite{jadhav2024designer}, optimization of truss structures has been performed by leveraging LLMs and feedback from results.
  In \cite{qiu2024robomorph}, a modular robot design optimization has been conducted based on sequential prompt adjustments using an LLM.
  A notable feature of LLMs is their ability to output appropriate designs based on language instructions when provided with problem settings, context, and various examples.
  However, while LLMs are effective for assisting with general design, they face challenges in fine-tuning detailed parameters.

  To address this, we propose a method that combines black-box optimization with LLMs to enable more efficient output of design solutions.
  Our approach leverages the strength of black-box optimization for precise numerical adjustments and the capability of LLMs to produce broad design outputs.
  In parallel with the sampling process in conventional black-box optimization, sampling is also performed using an LLM, which is provided with prior knowledge, problem settings, and design examples.
  It should be noted that while the combination of LLMs and BBO has been shown to be effective for single-objective and general numerical problems \cite{liu2024large}, this study extends the approach to more complex robot design problems and further expands it from single-objective to multi-objective optimization.
  In this study, a multi-objective robot design optimization is conducted by adjusting the robot's base position, joint configuration, and link lengths to minimize the end-effector position error and the required torque for given operation points.
  We demonstrate that our method enables more efficient exploration of design solutions compared to black-box optimization alone, while also discussing the method's interesting characteristics and limitations.

  The contributions of this study are as follows:
  \begin{itemize}
    \item We propose a robot body design optimization method that combines large language models with multi-objective black-box optimization.
    \item We demonstrate that the proposed method can explore design solutions more efficiently than black-box optimization alone.
  \end{itemize}

  The structure of this study is as follows.
  In \secref{sec:proposed}, we describe the problem setting, particularly the design parameters and objective functions.
  We also explain the sampling process using multi-objective black-box optimization and LLMs.
  In \secref{sec:experiment}, we present experimental results for three sets of target operation points.
  Several evaluation experiments are conducted, including examining changes in the results based on the proportion of LLM-based sampling and differences in prompts.
  Finally, we discuss the experimental results in \secref{sec:discussion} and present our conclusions in \secref{sec:conclusion}.
}%
{%
  これに対して, もう一つの選択肢として, 大規模言語モデル(LLM) \cite{achiam2023gpt4, kawaharazuka2024foundation}を用いた身体設計の例が出始めている.
  \cite{stella2023design}は大規模言語モデルがロボット設計の補助に有用であることを定性的に示している.
  \cite{jadhav2024designer}は大規模言語モデルと結果のフィードバックを用いたトラス構造の最適化を行った.
  \cite{qiu2024robomorph}は大規模言語モデルとプロンプトの逐次的な修正に基づき, モジュラー型ロボットの身体設計を行った.
  この大規模言語モデルは, 問題設定や状況, 様々な例示を与えることで, 言語指示に合わせた適切な設計を出力してくれることが魅力である.
  しかし, 大まかな設計の手助けには有効な一方で, 細かいパラメータの調整等には難がある.

  そこで本研究では, このブラックボックス最適化と大規模言語モデルを組み合わせることで, より効率的に設計解を出力可能な手法を開発する.
  ブラックボックス最適化の得意な細かい数値調整と大規模言語モデルによる大まかな設計出力の利点を活かした形である.
  通常のブラックボックス最適化によるサンプリングと並行し, 様々な事前知識や問題設定, 設計の例示を与えたLLMによるサンプリングを実行する.
  なお, LLMとBBOの併用は単目的かつ一般的な数値問題に対して有効であることが示されているが\cite{liu2024large}, 本研究はこれをより複雑なロボット設計に対して適用し, かつ単目的から多目的最適化に拡張したものと言える.
  本研究では与えられた動作点に対して, 手先の位置誤差と必要なトルクを最小化する評価関数を与え, ロボット原点と関節配置の順番, リンク長さを調整する多目的設計最適化を行う.
  本手法がブラックボックス最適化単体による最適化に比べ効率よく設計解を探索できることを示すと同時に, その興味深い特性や手法の限界についても議論する.

  本研究の貢献は以下である.
  \begin{itemize}
    \item 大規模言語モデルと多目的ブラックボックス最適化を組み合わせたロボット身体設計最適化手法の提案
    \item 提案手法がブラックボックス最適化単体に比べ効率的に設計解を探索できることの実証
  \end{itemize}

  本研究の構成は以下である.
  \secref{sec:proposed}では, 本研究の問題設定, 特に設計パラメータと評価関数について述べる.
  また, 多目的ブラックボックス最適化と大規模言語モデルを用いたサンプリングについて述べる.
  \secref{sec:experiment}では, 3つの指令動作点群に対する実験結果を述べる.
  このとき, LLMによるサンプリングの割合や, promptの違いに応じた結果の変化など, いくつかの評価実験を行う.
  最後に, \secref{sec:discussion}で実験結果について考察し, \secref{sec:conclusion}で結論を述べる.
}%

\section{Efficient Robot Design with Multi-Objective Black-Box Optimization and LLM-based Sampling} \label{sec:proposed}
\switchlanguage%
{%
  The problem setting of this study is briefly described (\figref{figure:overview}).
  Here, we focus on a robot arm with multiple joints connected in series.
  The robot is given target operation points $\bm{P}^{ref}_i$ ($1 \leq i \leq N_{ref}$, where $N_{ref}$ is the number of target operation points).
  The objective of this study is to find Pareto-optimal solutions that minimize both the position error $E^{pos}$, which measures how accurately the robot reaches the target points, and the total joint torque $E^{torque}$ required during the process.
  To achieve this objective, we utilize a combination of multi-objective black-box optimization (BBO) and large language models (LLMs).

  This study consists of the following four phases:
  \begin{itemize}
    \item Definition of design parameters and constraints (\secref{subsec:parameter})
    \item Computation of objective function values using inverse kinematics and torque evaluation (\secref{subsec:objective})
    \item Sampling with black-box optimization and a large language model at a fixed ratio (the first part of \secref{subsec:optimization} and \secref{subsec:llm})
    \item Evaluation of optimization progress using hypervolume (the latter part of \secref{subsec:optimization})
  \end{itemize}
}%
{%
  本研究の問題設定について簡潔に述べる(\figref{figure:overview}).
  ここでは複数の関節がシリアルにつながったロボットアームを題材として扱う.
  ロボットは到達すべき指令座標$\bm{P}^{ref}_i$ ($1 \leq i \leq N_{ref}$, $N_{ref}$は指令座標の数)が与えられる.
  これら指令座標にどの程度到達できるかの位置誤差$E^{pos}$, また, その際の関節トルクの合計$E^{torque}$を評価関数として, これらを最小化するようなパレート解を見つけるのが本研究の目的である.
  この目的に対して, 我々は多目的ブラックボックス最適化と大規模言語モデルを併用する.

  本研究は以下4つのフェーズから構成される.
  \begin{itemize}
    \item 設計パラメータと制約条件の定義(\secref{subsec:parameter})
    \item 逆運動学とトルク評価による評価関数値の算出(\secref{subsec:objective})
    \item ブラックボックス最適化と大規模言語モデルを用いたサンプリングを固定割合で実行(\secref{subsec:optimization}の前半と\secref{subsec:llm})
    \item ハイパーボリュームによる最適化の進捗評価(\secref{subsec:optimization}の後半)
  \end{itemize}
}%

\begin{figure}[t]
  \centering
  \includegraphics[width=0.98\columnwidth]{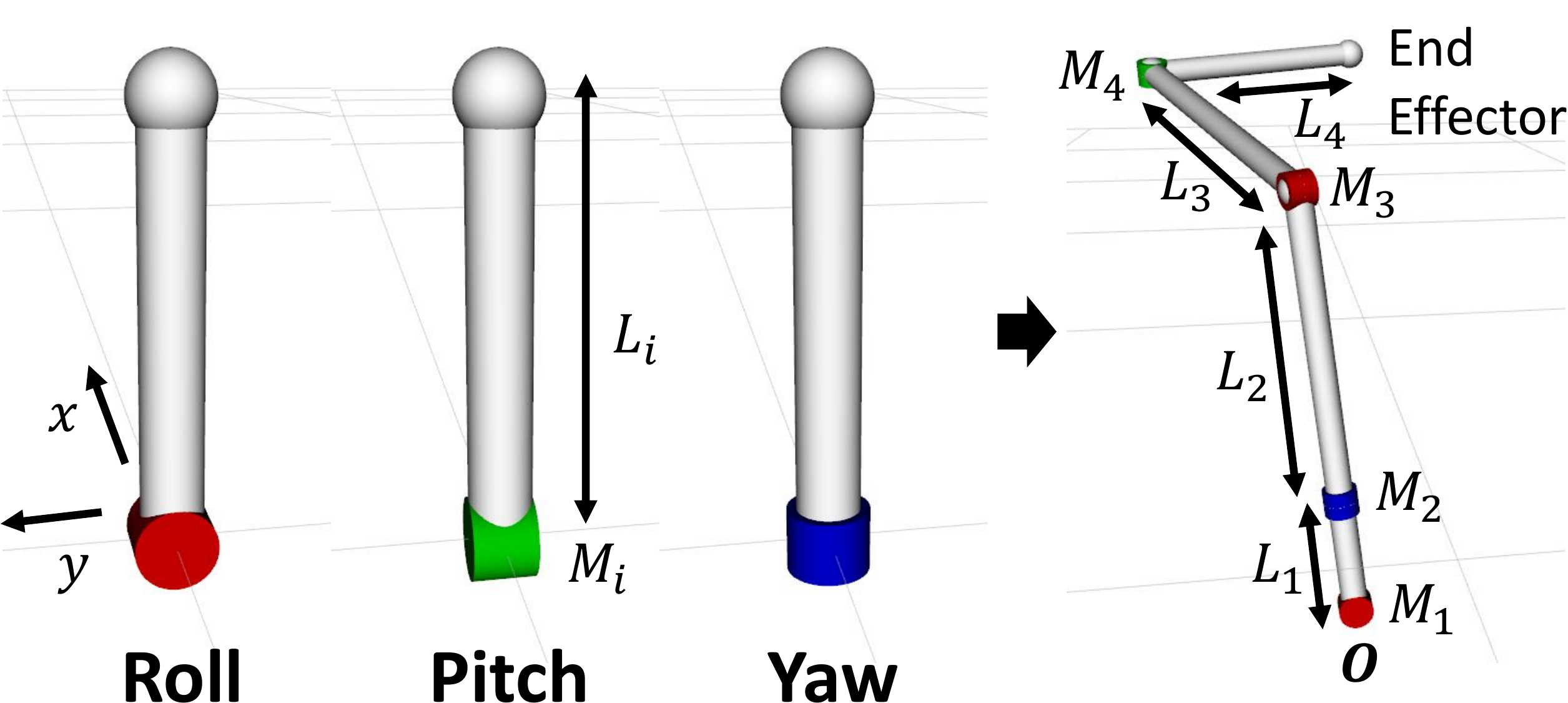}
  \vspace{-1.0ex}
  \caption{The robot design parameters handled in this study. The parameters include the base link coordinate $\bm{O}$, the joint types $\bm{M}$, and the link lengths $\bm{L}$ of the robot.}
  \vspace{-1.0ex}
  \label{figure:configuration}
\end{figure}

\begin{figure}[t]
  \centering
  \includegraphics[width=0.65\columnwidth]{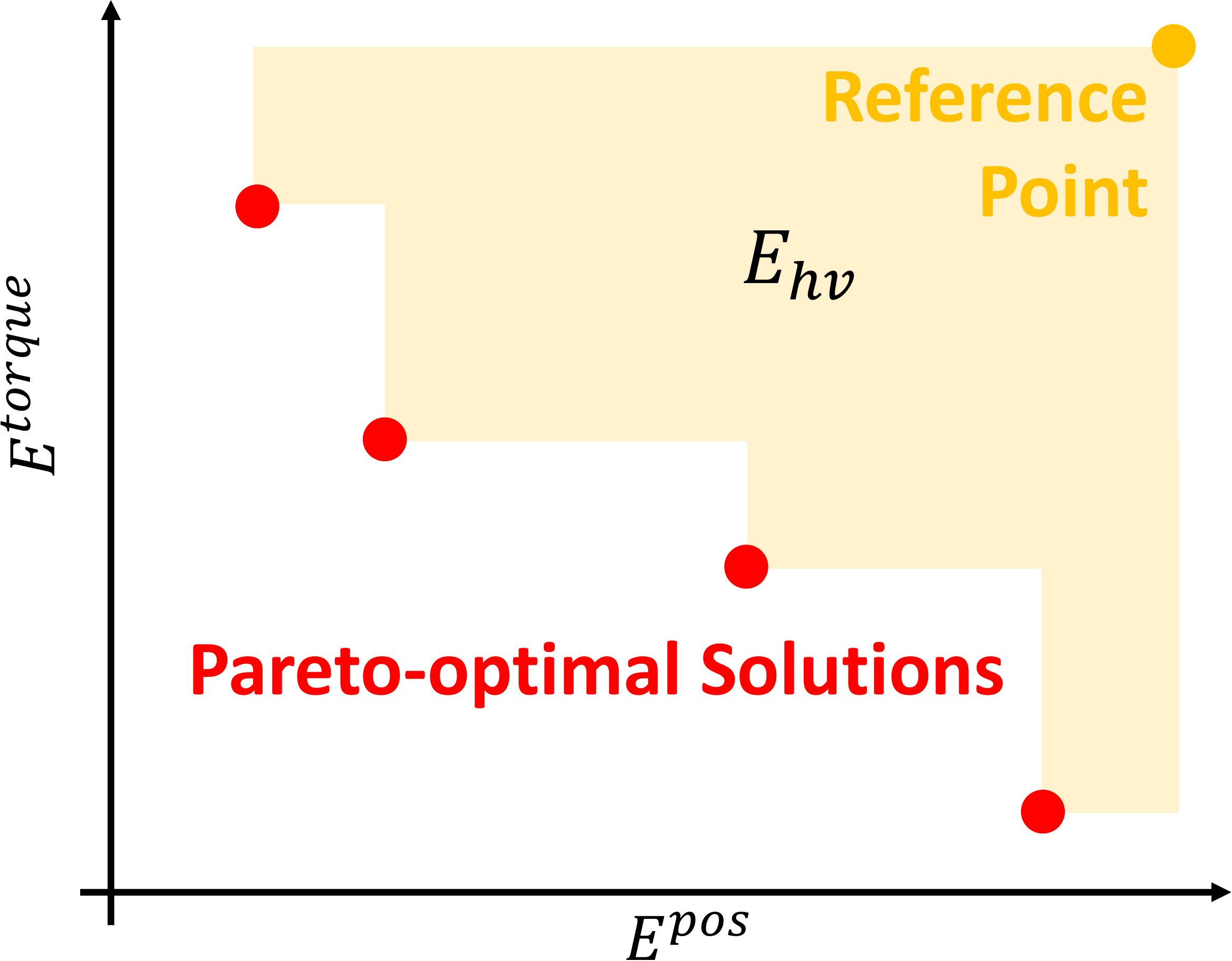}
  \vspace{-1.0ex}
  \caption{Hypervolume for the evaluation of multi-objective optimization.}
  \vspace{-1.0ex}
  \label{figure:evaluation}
\end{figure}

\begin{figure}[t]
  \centering
  \includegraphics[width=0.98\columnwidth]{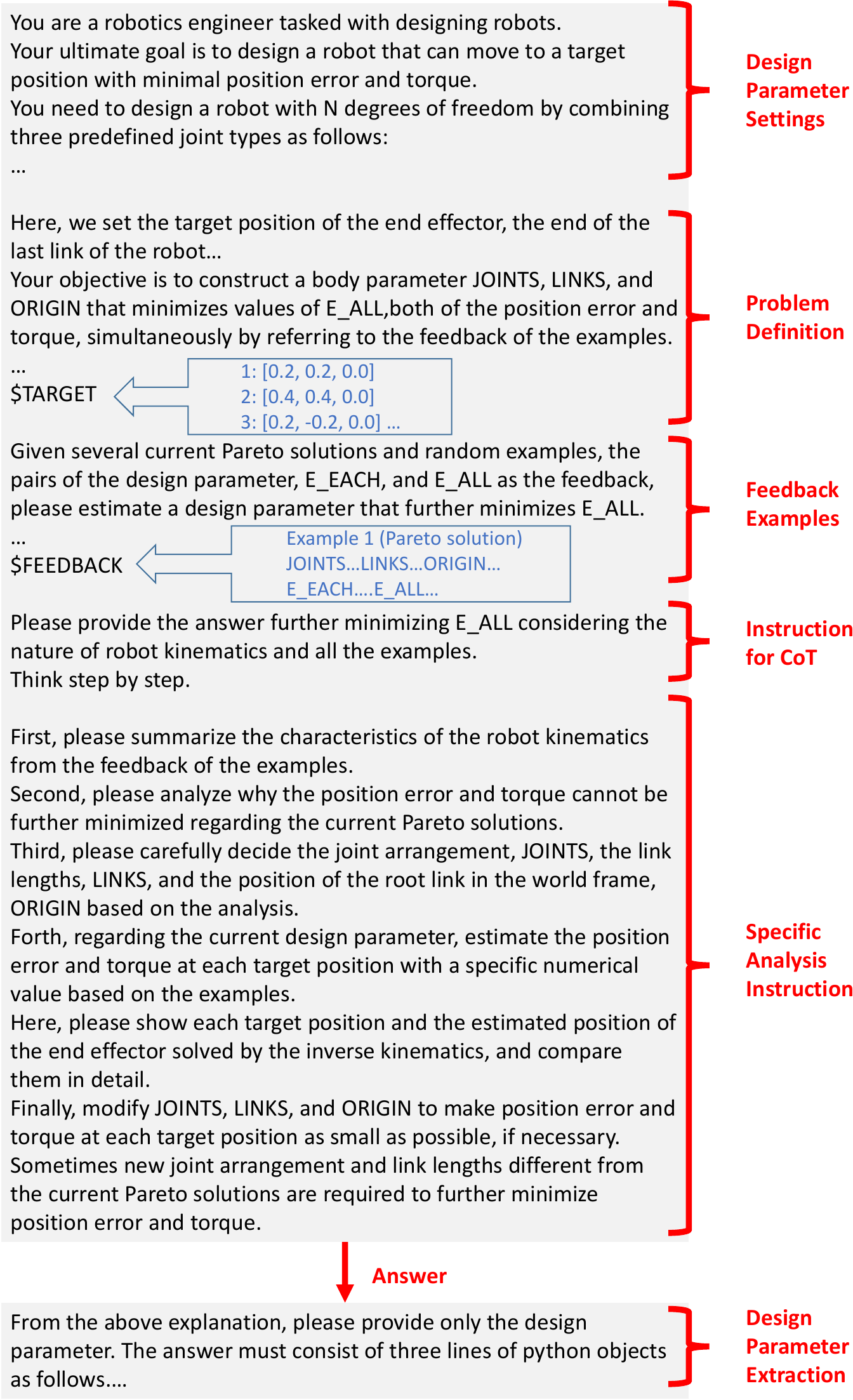}
  \vspace{-1.0ex}
  \caption{The overview of the prompt for LLM-based sampling conducted in this study.}
  \vspace{-1.0ex}
  \label{figure:prompt}
\end{figure}

\subsection{Robot Design Parameters} \label{subsec:parameter}
\switchlanguage%
{%
  The design parameters of the robot used in this study are shown in \figref{figure:configuration}.
  The robot consists of $D$ joints, and each joint $M_j$ ($1 \leq j \leq D$) is configured as either a Roll, Pitch, or Yaw joint.
  Each joint is followed by a link of length $L_j$.
  In this study, each joint can move within the range of $[-2.4, 2.4]$ rad, and the length of each link is set within the range of $[0.03, 0.3]$ m.
  When all joint angles are set to $0$, all links align vertically in a straight line.
  Additionally, the base link coordinate $\bm{O}$ of the robot in 3D space is also treated as a design parameter.
  This assumes that the robot is properly positioned for the task, particularly in systems such as mobile robots.
  The first joint $M_1$ of the robot is placed at coordinate $\bm{O}$.
  The values of $\bm{O}$ in the $x$, $y$, and $z$ directions are within the range $-1.0 \leq O_{x, y, z} \leq 1.0$ m.
  Thus, the full set of design parameters for the robot consists of $\bm{O}$, $\bm{M}$, and $\bm{L}$, forming a $2D+3$ dimensional vector that includes both discrete and continuous values.
  Once these parameters are determined, the corresponding Unified Robot Description Format (URDF) of the robot can be automatically generated.
}%
{%
  本研究で扱うロボットの設計パラメータについて述べる(\figref{figure:configuration}).
  ロボットは$D$個の関節からなり, 各関節$M_j$ ($1 \leq j \leq D$)はRoll, Pitch, Yaw関節のいずれかで構成される.
  各関節の後には長さ$L_j$のリンクが続く.
  本研究では, 各関節は$[-2.4, 2.4]$ radの範囲で動くことができ, リンクの長さは$[0.03, 0.3]$ mの範囲で設定される.
  全ての関節の角度が$0$であるとき, 全リンクは鉛直上向きに直線に並ぶ.
  加えて, ロボットの3次元空間上の原点座標$\bm{O}$も設計パラメータとして扱う.
  これは, タスクに対してロボットが適切な位置に配置されること, 特に台車型ロボットのような系を想定している.
  座標$\bm{O}$にロボットの1つ目の関節$M_1$が配置される.
  $O$のxyz方向の値は$-1.0 \leq O_{x, y, z} \leq 1.0$の範囲を取る.
  よって, ロボットの全設計パラメータは$\bm{O}$, $\bm{M}$, $\bm{L}$であり, 離散値と連続値を含む$2D+3$次元のベクトルとなる.
  これらのパラメータが決まると, それに対応するロボットのURDF (Unified Robot Description Format)を自動生成することができる.
}%

\begin{figure}[t]
  \centering
  \includegraphics[width=0.98\columnwidth]{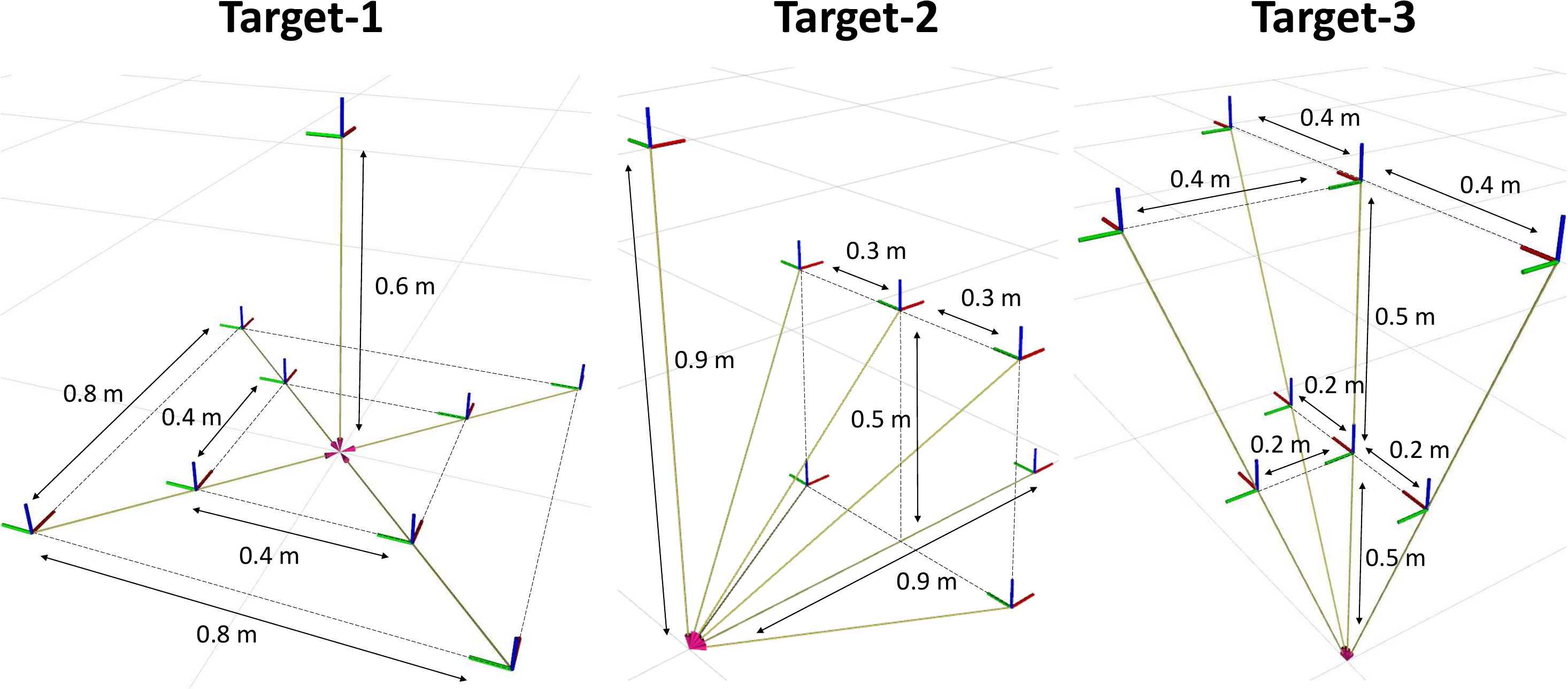}
  \vspace{-1.0ex}
  \caption{Experimental setup of the three target operation points: Target-1, Target-2, and Target-3.}
  \vspace{-1.0ex}
  \label{figure:exp-setup}
\end{figure}

\subsection{Objective Functions} \label{subsec:objective}
\switchlanguage%
{%
  The objective functions for robot design optimization can take various forms; however, in this study, we use a relatively simple formulation consisting of the position error $E^{pos}$ with respect to the target positions and the joint torque $E^{torque}$.
  These functions are defined as follows.
  \begin{align}
    E^{pos} &= \sum^{N_{ref}}_{i} ||\bm{P}^{ik}_i - \bm{P}^{ref}_i||_{2}\\
    E^{torque} &= \alpha\sum^{N_{ref}}_{i} ||\bm{\tau}^{ik}_i||_{2}
  \end{align}
  where $\bm{P}^{ik}_i$ and $\bm{\tau}^{ik}_i$ represent the end-effector position and gravity-compensation torque, respectively, when the robot solves the inverse kinematics (IK) to reach the target position $\bm{P}^{ref}_i$, assuming the robot's initial posture has all joint angles set to zero.
  The inverse kinematics algorithm used in this study is based on the methods described in \cite{chan1995ik, sugiura2007ik}.
  Joint angle limits are considered, whereas collision avoidance is not taken into account.
  The position error is defined as the minimum distance between the end effector and the target position achieved during 100 iterations of the inverse kinematics solver.
  Joint torques are computed by assuming each link to be a rigid body with uniform density.
  In other words, this evaluates whether the target end-effector position can be achieved with that body design, and how much joint torque is required in doing so.
  Of course, it is also possible to consider not only the target position but also the orientation, or to use a more complex evaluation function.

  Note that, for simplicity, the posture of the target points is ignored, and only the reachability to the target positions is considered.
  Additionally, the coefficient $\alpha$ (set to $\alpha=40$ in this study) is multiplied by $E^{torque}$ to scale it to a similar magnitude as $E^{pos}$ for clarity.
}%
{%
  身体設計最適化の評価関数は多様な形式が可能であるが, 本研究では比較的シンプルな形として指令位置に対する位置誤差$E^{pos}$と関節トルク$E^{torque}$を扱う.
  これらは以下のように定義される.
  \begin{align}
    E^{pos} &= \sum^{N_{ref}}_{i} ||\bm{P}^{ik}_i - \bm{P}^{ref}_i||_{2}\\
    E^{torque} &= \alpha\sum^{N_{ref}}_{i} ||\bm{\tau}^{ik}_i||_{2}
  \end{align}
  ここで, $\bm{P}^{ik}_i$と$\bm{\tau}^{ik}_i$はロボットが$\bm{P}^{ref}_i$に対して逆運動学(IK)を解いたときの, 手先位置とその際の重力補償トルクである.
  なお, 逆運動学のアルゴリズムは\cite{chan1995ik, sugiura2007ik}におけるアルゴリズムを用いており, 関節角度制限は考慮しているが, 衝突回避等は考慮していない.
  手先誤差は逆運動学の100イテレーションの間に最大で手先が指令位置に近づいた際の誤差を用い, トルクは各リンクを密度一定の剛体と仮定して算出している.
  つまり, その身体設計で, 指令手先位置が実現できるか, その際にどのくらいの関節トルクが必要かを評価している.
  もちろん指令位置だけでなく姿勢を考慮したり, より複雑な評価関数を用いることも可能である.
  また, $E^{torque}$に係数$\alpha$ (本研究では$\alpha=40$)をかけ, 分かりやすいよう$E^{pos}$と同程度のスケールに調整している.
}%

\subsection{Multi-Objective Black-Box Optimization} \label{subsec:optimization}
\switchlanguage%
{%
  In this study, we address a multi-objective optimization problem that aims to minimize both objective functions $E^{pos}$ and $E^{torque}$.
  This allows us to obtain various design solutions from the Pareto-optimal set.
  For the multi-objective optimization, we use the Tree-Structured Parzen Estimator (TPE) \cite{bergstra2011tpe} implemented in the black-box optimization library Optuna \cite{akiba2019optuna}.
  TPE supports multi-objective optimization and is notable for requiring a relatively small recommended sample size of fewer than 1000.
  Non-dominated Sorting Genetic Algorithm II (NSGA-II) \cite{deb2002nsgaii} is also commonly used; however, it typically requires a larger number of samples and is less suitable than TPE for handling the integer and continuous parameters that frequently appear in design optimization problems.

Since comparing BBO algorithms is not the primary objective of this study, we do not include such comparisons in this work.

  To evaluate the performance of the multi-objective optimization, we use hypervolume \cite{zitzler2001spea2} as a metric to measure how effectively good design solutions are explored.
  As shown in \figref{figure:evaluation}, the hypervolume $E_{hv}$ is defined as the area enclosed by the Pareto-optimal solutions (represented by red points) and the reference point.
  A larger hypervolume indicates that more favorable design solutions have been explored.
  In this study, the reference point is set to $(5.0, 5.0)$.
}%
{%
  本研究では, 評価関数$E^{pos}$と$E^{torque}$の両方を最小化する多目的最適化問題を扱う.
  これによって, パレート解から様々な設計解を得ることができる.
  本研究の多目的最適化には, Black-Box OptimizationのライブラリであるOptuna \cite{akiba2019optuna}に実装されたTree-Structured Parzen Estimator (TPE) \cite{bergstra2011tpe}を用いる.
  このTPEは多目的最適化が可能かつ推奨サンプリング数が1000以下と比較的少ないという特徴がある.
  Non-dominated Sorting Genetic Algorithms II (NSGA-II) \cite{deb2002nsgaii}もよく使われるが, より多くのサンプル数が必要であり, かつTPEに比べると設計最適化に多い整数や小数の扱いに不向きである.
  なお, BBOアルゴリズムの比較が主目的ではないため, 本研究ではそれらの比較は行わない.

  この多目的最適化を評価するにあたり, 多目的最適化がどれだけ良い設計解を探索出来ているかを表す評価指標として, Hypervolume \cite{zitzler2001spea2}を用いる.
  これは, \figref{figure:evaluation}に示すように, 赤点で示したパレート解とReference Pointによって囲まれた領域の面積$E_{hv}$として定義される.
  この値が大きいほど, より多くの良い設計解を探索できている.
  なお, 本研究ではReference Pointを$(5.0, 5.0)$として設定した.
}%

\begin{figure*}[t]
  \centering
  \includegraphics[width=2.0\columnwidth]{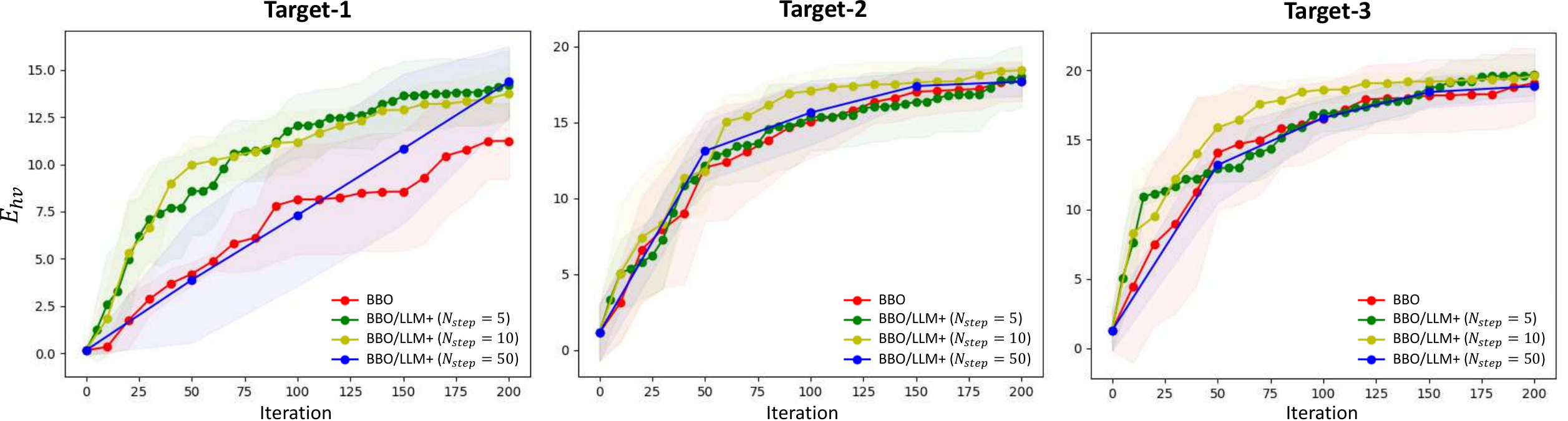}
  \vspace{-1.0ex}
  \caption{Comparison of \textbf{BBO}, \textbf{BBO/LLM+} ($N_{step}=5$), \textbf{BBO/LLM+} ($N_{step}=10$), and \textbf{BBO/LLM+} ($N_{step}=50$) for Target-1, Target-2, and Target-3.}
  \vspace{-1.0ex}
  \label{figure:exp-ablation-1}
\end{figure*}

\begin{figure*}[t]
  \centering
  \includegraphics[width=2.0\columnwidth]{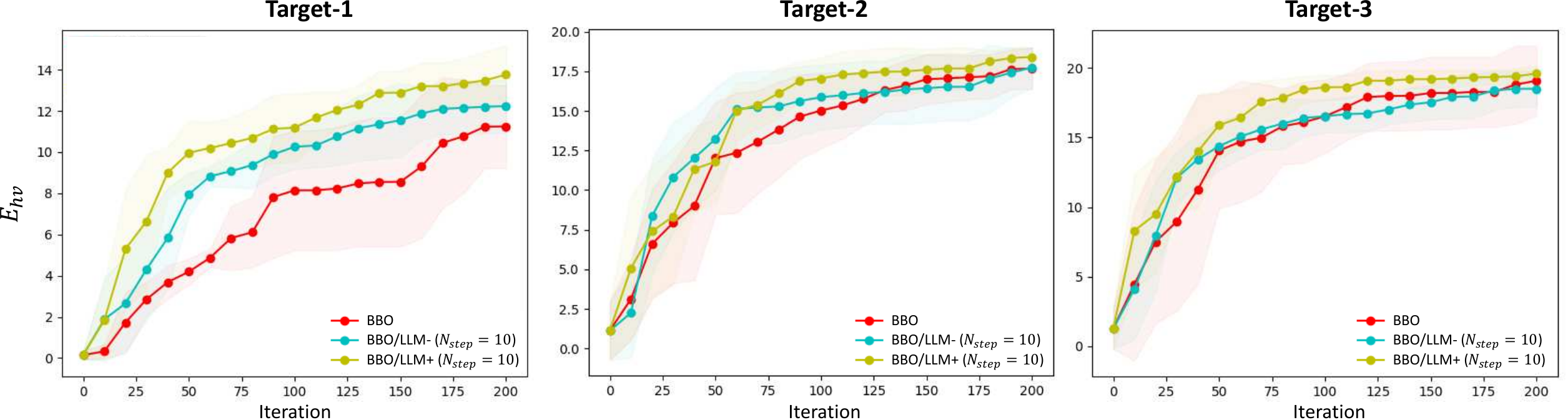}
  \vspace{-1.0ex}
  \caption{Comparison of \textbf{BBO}, \textbf{BBO/LLM-} ($N_{step}=10$), and \textbf{BBO/LLM+} ($N_{step}=10$) for Target-1, Target-2, and Target-3.}
  \vspace{-1.0ex}
  \label{figure:exp-ablation-2}
\end{figure*}

\subsection{LLM-based Sampling} \label{subsec:llm}
\switchlanguage%
{%
  In parallel with the multi-objective optimization sampling described above, sampling using a large language model is also performed.
  The LLM utilizes In-context Learning \cite{dong2022survey} to output the next design parameters to explore based on the given problem setting and feedback from previously obtained solutions.
  The actual prompt, feedback on the solutions, and the subsequent output formatting steps are illustrated in \figref{figure:prompt}.

  First, as shown at the top of \figref{figure:prompt}, the design parameters and problem setting of this study are described, and the target position $\bm{P}^{ref}_{i}$ is input into \$TARGET.
  Next, as feedback, $N_{pareto}$ solutions are randomly selected from the Pareto-optimal solutions and $N_{random}$ solutions are selected from the overall sampling results.
  The information for each solution includes the ORIGIN $\bm{O}$, JOINTS $\bm{M}$, and LINKS $\bm{L}$, the evaluation values E\_EACH $(E^{pos}_{i}, E^{torque}_{i})$ for each $\bm{P}^{ref}_{i}$, the end-effector positions $\bm{P}^{ik}_{i}$ and the joint torques $\bm{\tau}^{ik}_{i}$ obtained from the inverse kinematics solution, and the overall evaluation values E\_ALL $(E^{pos}, E^{torque})$.
  Instructions for Chain-of-Thought (CoT) reasoning, such as ``Think step by step'' \cite{kojima2022large}, and additional instructions on how to analyze specific parameters are provided.
  When only general CoT instructions are given, we refer to the approach as \textbf{LLM-}, while the approach that includes specific parameter analysis instructions tailored to this study is referred to as \textbf{LLM+}.
  The output obtained from this process is reformatted by the LLM to provide design parameters in the form of a Python List or Tuple.
  The generated solutions are then added to the sampling process of the multi-objective optimization.

  The optimization process starts with $N_{init}$ random sampling steps.
  After that, the optimization proceeds in steps of $N_{step}$, where one sampling step is performed using the LLM, followed by $N_{step}-1$ sampling steps using multi-objective black-box optimization.
  Although these processes can be executed in parallel, this sequential approach is adopted for ease of evaluation.
  The smaller the $N_{step}$ is, the greater the influence of the LLM is.
  By varying $N_{step}$, it is possible to evaluate the degree to which the LLM contributes to the optimization.
  Although $N_{init}$ is typically included in the total optimization attempts, for clarity, we reset the count to 0 after the $N_{init}$ random trials and perform a total of $N_{total}$ sampling steps.
}%
{%
  上記で説明した多目的最適化によるサンプリングと並行して, 大規模言語モデルを用いたサンプリングを行う.
  大規模言語モデルはIn-context Learning \cite{dong2022survey}を利用して, 与えられた問題設定やこれまで得られた解のフィードバックを元に, 次に探索すべき設計パラメータを出力する.
  実際のプロンプトや解のフィードバック, その後の出力成形の手順を\figref{figure:prompt}に示す.
  まず\figref{figure:prompt}の上部に示すように本研究の設計パラメータと問題設定を記述し, \$TARGETとして$\bm{P}^{ref}_{i}$を入力する.
  次にフィードバックとして, これまで得られたパレート解からランダムに$N_{pareto}$個, 全体のサンプリングから$N_{random}$個の解を入力する.
  ここでは, 各解のORIGIN $\bm{O}$, JOINTS $\bm{M}$, LINKS $\bm{L}$の情報, 各$\bm{P}^{ref}_{i}$に対する評価値E\_EACH $(E^{pos}_{i}, E^{torque}_{i})$, 逆運動学を解いた際の$\bm{P}^{ik}_{i}$と$\bm{\tau}^{ik}_{i}$, 全体の評価値E\_ALL $(E^{pos}, E^{torque})$を記述する.
  そしてChain-of-Thought (CoT)のための指示``Think step by step'' \cite{kojima2022large}と, 具体的なパラメータの分析方法に関する指示を加える
  なお, 一般的なCoTまでの指示を行う場合を\textbf{LLM-}, 本研究に特有な具体的なパラメータ分析指示を加える場合を\textbf{LLM+}と呼び区別する.
  ここで得られた出力から, 再度LLMを用いて設計パラメータをPythonのListまたはTupleの形で出力する.
  得られた解を多目的最適化のサンプリングに加え, 評価を行う.

  最適化では最初に$N_{init}$回のランダムサンプリングが実行される.
  その後, 最適化は$N_{step}$を区切りとして進行し, 大規模言語モデルを用いたサンプリングを1回行ったあと, 多目的ブラックボックス最適化によるサンプリングを$N_{step}-1$回行う.
  実際にはこれらを並行して実行できるが, 本研究では評価しやすいようにこのような形を取っている.
  $N_{step}$が小さいほどLLMの影響力は大きくなり, この$N_{step}$を変化させることで, どの程度LLMが最適化に寄与するかを評価することができる.
  一般的には$N_{init}$も最適化の試行に含まれるが, ここでは分かりやすいように$N_{init}$回のランダム試行が終わったところを0として, $N_{total}$回のサンプリングを行う.
}%

\section{Experiments} \label{sec:experiment}

\subsection{Experimental Setup} \label{subsec:exp-setup}
\switchlanguage%
{%
  In this study, experiments are conducted for three sets of target operation points, Target-1, Target-2, and Target-3, as shown in \figref{figure:exp-setup}.
  For each target operation point, we compare \textbf{BBO}, which uses only black-box optimization, and \textbf{BBO/LLM+}, which combines black-box optimization with a large language model.
  To evaluate the extent of LLM's contribution to the optimization process, we vary $N_{step}$, which represents the number of black-box optimization steps performed between each LLM-based sampling.
  In this study, experiments are conducted with $N_{step}={5, 10, 50}$.
  Additionally, we perform a comparative evaluation of \textbf{BBO/LLM-}, which uses general Chain-of-Thought reasoning without specific parameter analysis instructions.
  The other parameters are set as follows: $D=4$, $N_{init}=10$, $N_{pareto}=5$, $N_{random}=5$, and $N_{total}=200$.
  GPT-4o was used as the LLM, with the temperature set to 0.
  Each experiment is repeated five times with different random seed values, and the mean and standard deviation of the results are reported.
}%
{%
  本研究では, \figref{figure:exp-setup}に示す3つの指令動作点Target-1, Target-2, Target-3に対する実験を行う.
  各指令動作点に対して, ブラックボックス最適化のみを適用した\textbf{BBO}と大規模言語モデルを併用した\textbf{BBO/LLM+}を比較する.
  このとき, 何回のブラックボックス最適化に対して一回LLM-basedサンプリングを行うかを表す$N_{step}$を変化させることで, どの程度LLMが最適化に寄与するかを評価する.
  本研究では, $N_{step}=\{5, 10, 50\}$と変化させて実験を行う.
  また, 具体的なパラメータ分析指示を加えず一般的なCoTを用いる\textbf{BBO/LLM-}についても比較評価する.
  その他のパラメータについては, $D=4$, $N_{init}=10$, $N_{pareto}=5$, $N_{random}=5$, $N_{total}=200$と設定して実験を行った.
  また, LLMはGPT-4をtemperatureを0にして使用した.
  各実験はrandom seedの値を変えながら5回行い, その平均と標準偏差を表示している.
}%

\begin{table}[htb]
  \centering
  \caption{Comparison of hypervolume $E_{hv}$ and its standard deviation $\sigma_{hv}$ for \textbf{BBO}, \textbf{BBO/LLM-}, and \textbf{BBO/LLM+}}.
  \begin{tabular}{l||c|c|c|c|c|c}
    Method & \multicolumn{2}{|c|}{Target-1} & \multicolumn{2}{|c|}{Target-2} & \multicolumn{2}{|c}{Target-3} \\ \hline
    & $E_{hv}\uparrow$ & $\sigma_{hv}\downarrow$ & $E_{hv}\uparrow$ & $\sigma_{hv}\downarrow$ & $E_{hv}\uparrow$ & $\sigma_{hv}\downarrow$ \\ \hline\hline
    \textbf{BBO}      &  6.70 & 2.03 & 13.16 & 2.39 & 14.64 & 3.43 \\
    \textbf{BBO/LLM-} &  8.85 & \textbf{1.38} & 13.87 & 2.33 & 14.73 & 1.67 \\
    \textbf{BBO/LLM+} & \textbf{10.25} & 1.42 & \textbf{14.43} & \textbf{1.89} & \textbf{16.29} & \textbf{1.53} \\
  \end{tabular}
  \label{table:exp-ablation-2}
\end{table}

\begin{figure*}[t]
  \centering
  \includegraphics[width=2.0\columnwidth]{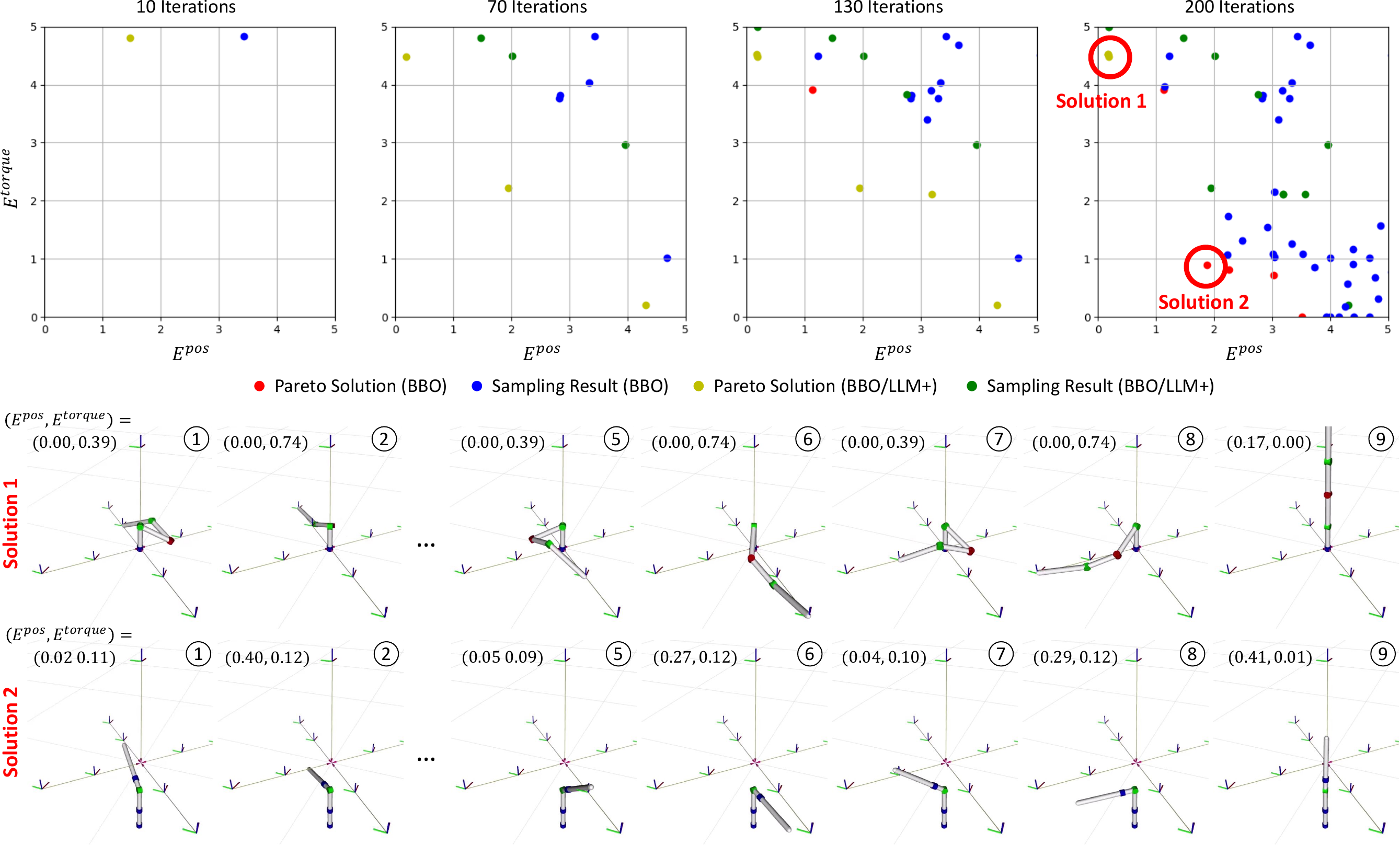}
  \vspace{-1.0ex}
  \caption{The sampling results of Target-1 for \textbf{BBO/LLM+} ($N_{step}=10$) and the selected Pareto-optimal solutions.}
  \vspace{-1.0ex}
  \label{figure:exp-target-1}
\end{figure*}

\begin{figure}[t]
  \centering
  \includegraphics[width=0.98\columnwidth]{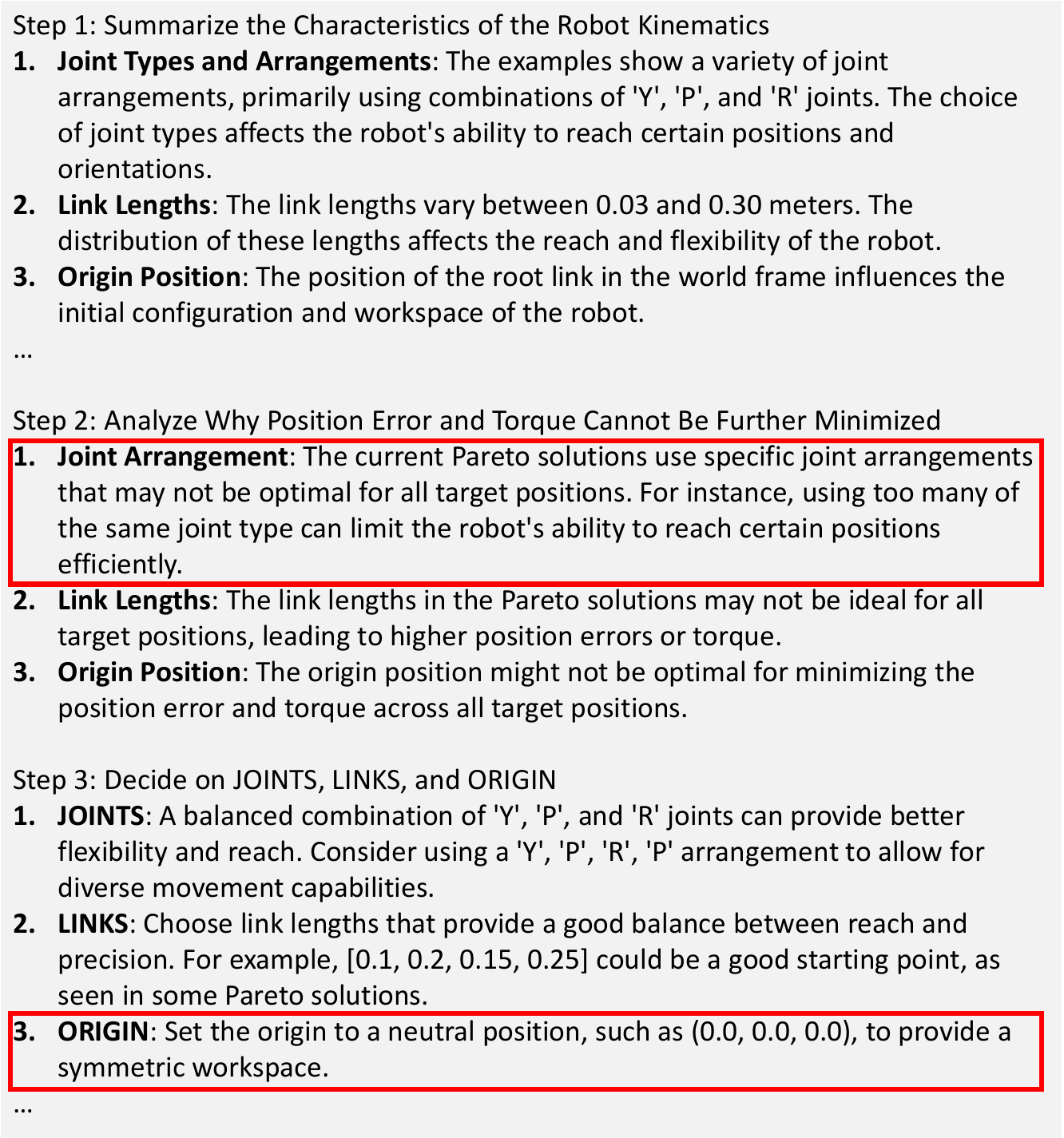}
  \vspace{-1.0ex}
  \caption{An overview of LLM output for Solution 1 in Target-1 experiment.}
  \vspace{-1.0ex}
  \label{figure:exp-target-1-llm}
\end{figure}

\begin{figure*}[t]
  \centering
  \includegraphics[width=2.0\columnwidth]{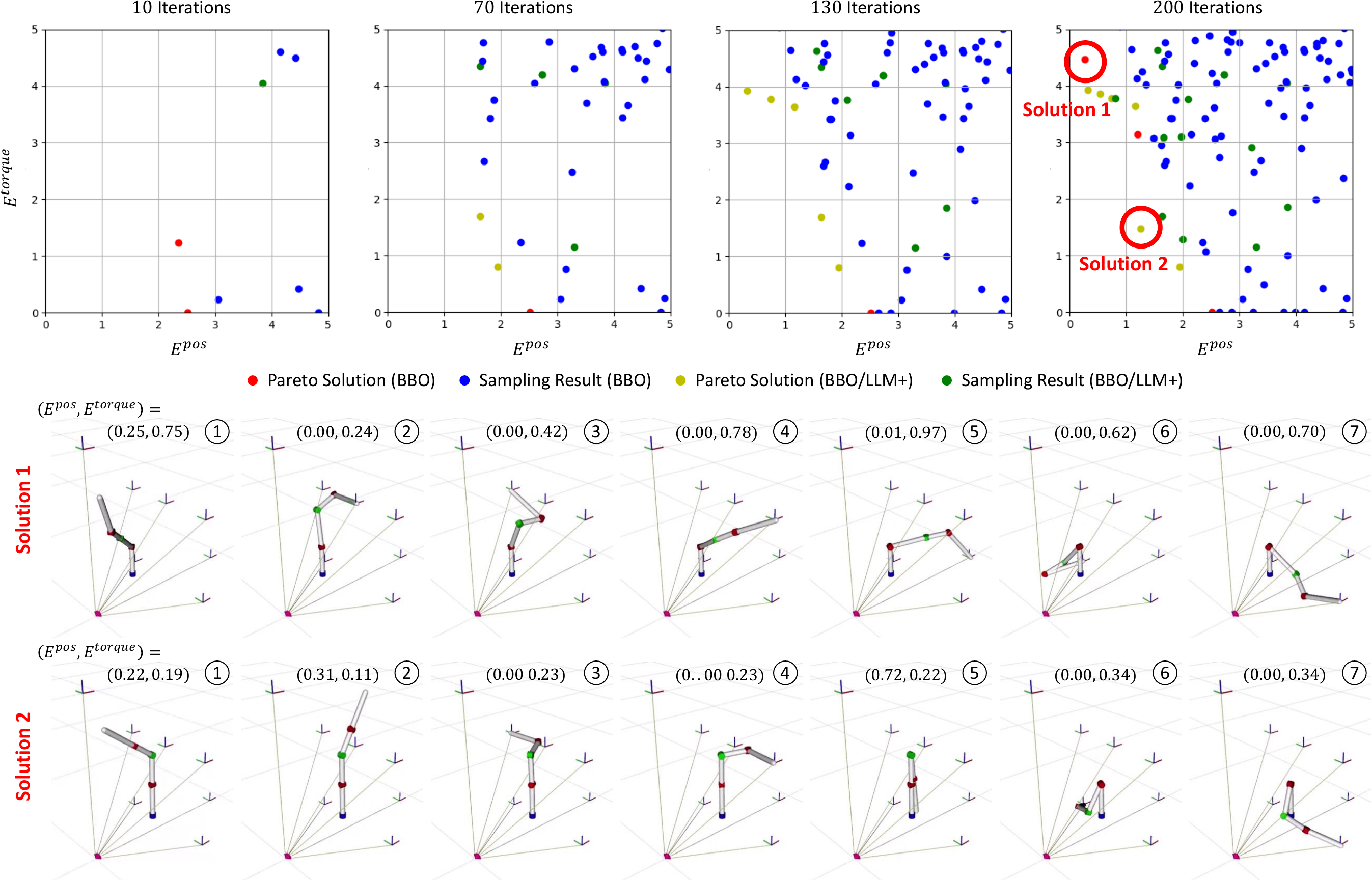}
  \vspace{-1.0ex}
  \caption{The sampling results of Target-2 for \textbf{BBO/LLM+} ($N_{step}=10$) and the selected Pareto-optimal solutions.}
  \vspace{-1.0ex}
  \label{figure:exp-target-2}
\end{figure*}

\begin{figure*}[t]
  \centering
  \includegraphics[width=2.0\columnwidth]{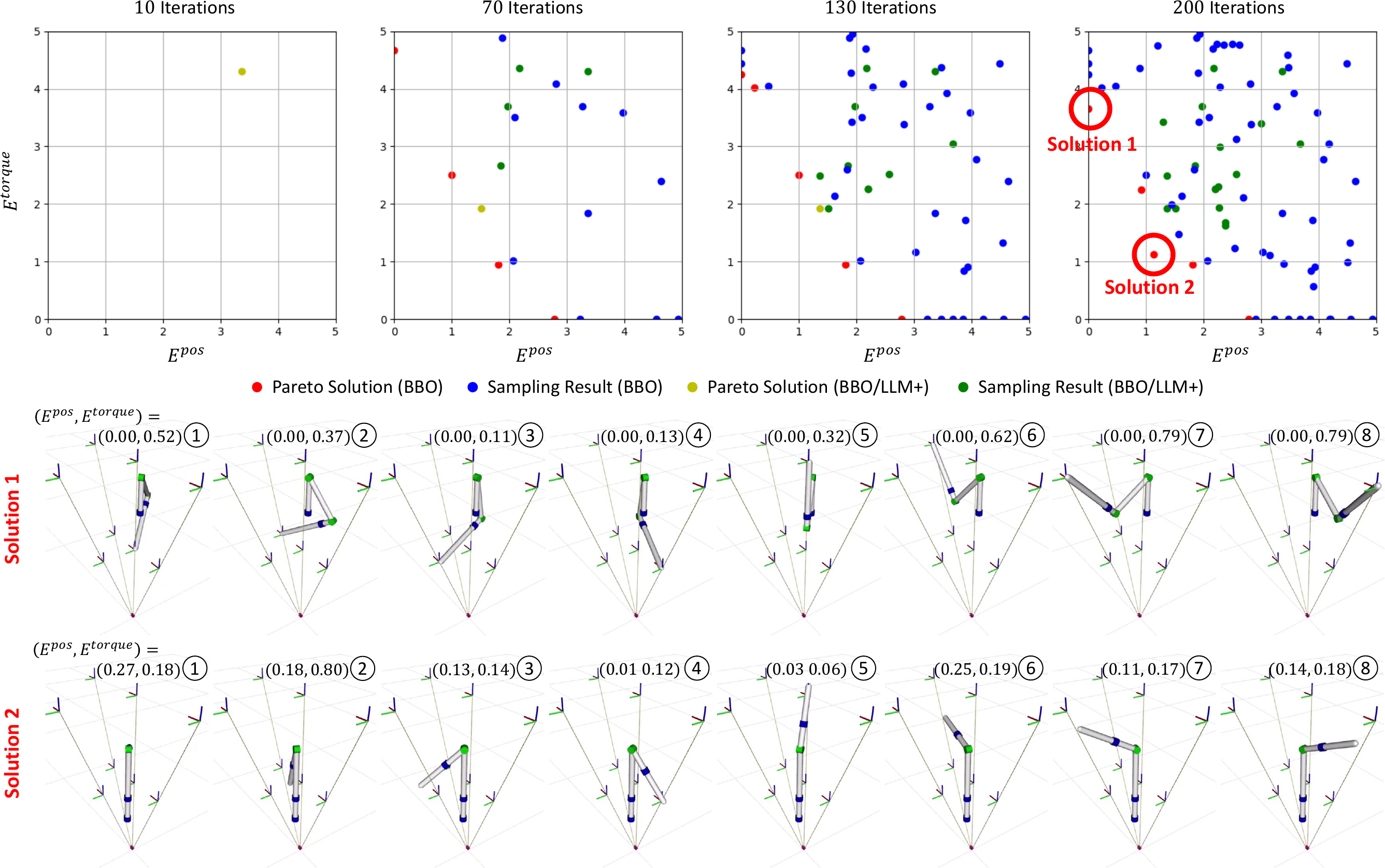}
  \vspace{-1.0ex}
  \caption{The sampling results of Target-3 for \textbf{BBO/LLM+} ($N_{step}=10$) and the selected Pareto-optimal solutions.}
  \vspace{-1.0ex}
  \label{figure:exp-target-3}
\end{figure*}

\subsection{Experimental Results} \label{subsec:exp-results}
\switchlanguage%
{%
  The experimental results are presented below.
  \figref{figure:exp-ablation-1} shows a comparison of the methods \textbf{BBO}, \textbf{BBO/LLM+} ($N_{step}=5$), \textbf{BBO/LLM+} ($N_{step}=10$), and \textbf{BBO/LLM+} ($N_{step}=50$) for Target-1, Target-2, and Target-3.
  The horizontal axis represents the optimization iterations, and the vertical axis represents the hypervolume $E_{hv}$.
  The figure illustrates how $E_{hv}$ changes with different values of $N_{step}$.
  A general trend observed is that \textbf{BBO/LLM+} ($N_{step}=10$) consistently demonstrates the best results across all target operation points, confirming the effectiveness of LLM-based sampling.
  For Target-1, the improvement is particularly significant, while for Target-2 and Target-3, the impact is less pronounced, indicating that the effectiveness of LLM-based sampling depends on the target operation points.
  For $N_{step}=50$, only four LLM-based samplings occur within 200 sampling steps, making the trend of $E_{hv}$ similar to that of \textbf{BBO}.
  For $N_{step}=5$, although LLM-based sampling occurs more frequently and is effective in the early stages for Target-1 and Target-2, the results are inferior compared to $N_{step}=10$.

  \figref{figure:exp-ablation-2} shows a comparison of \textbf{BBO}, \textbf{BBO/LLM-} ($N_{step}=10$), and \textbf{BBO/LLM+} ($N_{step}=10$).
  The general trend indicates that \textbf{BBO/LLM-} performs better than \textbf{BBO}, but it is inferior to \textbf{BBO/LLM+}, demonstrating the importance of adding specific parameter analysis instructions.
  The average hypervolume $E_{hv}$ over all iterations and the average standard deviation $\sigma_{hv}$ of $E_{hv}$ are shown in \tabref{table:exp-ablation-2}.
  The numerical results confirm the performance trend of \textbf{BBO} $<$ \textbf{BBO/LLM-} $<$ \textbf{BBO/LLM+}.
  Moreover, the values of $\sigma_{hv}$ for \textbf{BBO/LLM-} and \textbf{BBO/LLM+} are smaller than those for \textbf{BBO}, indicating that LLM-based sampling reduces randomness and leads to more stable results.
}%
{%
  実験結果について述べる.
  \figref{figure:exp-ablation-1}にはTarget-1, Target-2, Target-3に対して\textbf{BBO}, \textbf{BBO/LLM+} ($N_{step}=5$), \textbf{BBO/LLM+} ($N_{step}=10$), \textbf{BBO/LLM+} ($N_{step}=50$)を適用した際の比較結果を示す.
  横軸を最適化のイテレーション, 縦軸をHypervolume $E_{hv}$としており, $N_{step}$ごとに$E_{hv}$がどのように変化するかを示している.
  全体的な傾向として, 特に\textbf{BBO/LLM+} ($N_{step}=10$)が最も良い結果を示しており, どの動作指令点に対してもLLMによるサンプリングが効果的であることがわかる.
  Target-1に関しては効果は絶大だが, Target-2, Target-3に関してその効果は劣り, 指令動作点によって効果が異なることがわかる.
  $N_{step}=50$の場合は200回のサンプリングのうち4回しかLLMによるサンプリングが行われていないため, $E_{hv}$の傾向は\textbf{BBO}とほぼ同様である.
  $N_{step}=5$の場合はLLMによるサンプリング回数がかなり多くなり, Target-1とTarget-2の初期では効果的であるが, $N_{step}=10$の場合に比べると結果は劣る.

  次に, \figref{figure:exp-ablation-2}には\textbf{BBO}, \textbf{BBO/LLM-} ($N_{step}=10$), \textbf{BBO/LLM+} ($N_{step}=10$)を適用した際の比較結果を示す.
  全体的な傾向として, \textbf{BBO/LLM-}は\textbf{BBO}に比べて若干良い結果を示しているが, \textbf{BBO/LLM+}に比べると劣り, 具体的なパラメータ分析指示を加えることが重要であることがわかる.
  この\figref{figure:exp-ablation-2}の結果について, 全イテレーションにおけるHypervolume $E_{hv}$の平均値と, $E_{hv}$の標準偏差$\sigma_{hv}$の平均値を\tabref{table:exp-ablation-2}に示す.
  数値として見ても, \textbf{BBO}$<$\textbf{BBO/LLM-}$<$\textbf{BBO/LLM+}という同様の性能の傾向が見られる.
  また, $\sigma_{hv}$の値は\textbf{BBO/LLM-}や\textbf{BBO/LLM+}の方が\textbf{BBO}よりも小さいことがわかる.
  $\sigma_{hv}$は結果の安定性を示しているが, LLMのサンプリングによりランダム性が減少し, 安定した結果が得られることがわかる.
}%

\subsection{Target-1 Experiment} \label{subsec:exp-1}
\switchlanguage%
{%
  We now discuss the experimental results for each target operation point, focusing on \textbf{BBO/LLM+} ($N_{step}=10$).
  First, we explain how to interpret the figures using \figref{figure:exp-target-1} as an example.
  The top section of the figure shows the sampling results after 10, 70, 130, and 200 iterations, where each point represents a sampling result obtained from either BBO or LLM.
  The horizontal axis represents $E^{pos}$, and the vertical axis represents $E^{torque}$, with Pareto-optimal solutions displayed in a different color.
  Two Pareto-optimal solutions, Solution 1 and Solution 2, are selected from the results after 200 iterations, and the designs and inverse kinematics results for each target operation point are shown in the bottom section of the figure.

  We describe the experimental results for Target-1 (\figref{figure:exp-target-1}).
  In the case of Target-1, many of the Pareto-optimal solutions across iterations are obtained from LLM-based sampling.
  Solution 1 represents a solution with low $E^{pos}$ but high $E^{torque}$, while Solution 2 is the opposite, featuring low $E^{torque}$ but higher $E^{pos}$.

  Solution 1 places $\bm{O}$ at the origin in 3D space, providing easy access to all target operation points, and adopts a joint arrangement of $Y-P-R-P$ (where $R$ represents Roll, $P$ represents Pitch, and $Y$ represents Yaw), achieving uniform reachability across all operation points.
  In contrast, Solution 2 shifts $\bm{O}$ in the negative $z$-axis direction and adopts a joint arrangement of $Y-Y-P-Y$, which reduces reachability to the target operation points but minimizes the required torque.

  As an example, \figref{figure:exp-target-1-llm} shows a portion of the LLM output when Solution 1 was sampled.
  In Step 1, the characteristics of each parameter are analyzed.
  In Step 2, the current issues with the Pareto-optimal solutions are examined.
  In Step 3, the LLM outputs the new set of design parameters.
  In the analysis of the joint arrangement in Step 2, many of the previous solutions used multiple joints of the same type consecutively, but the LLM explicitly changed this to the arrangement $Y-P-R-P$.
  Additionally, in Step 3, the base link coordinate $\bm{O}$ is set precisely at the origin to secure a symmetric workspace.
  The ability to secure a symmetric workspace is particularly important for Target-1, as placing $\bm{O}$ at the origin significantly simplifies the problem.
  This type of reasoning is difficult for black-box optimization alone.
  However, the LLM can determine an appropriate $\bm{O}$ based on the problem setting, demonstrating the effectiveness of LLM-based sampling in achieving better design solutions.
}%
{%
  ここからは, 各動作指令点に関して個別に\textbf{BBO/LLM+} ($N_{step}=10$)の際の実験結果を述べる.
  まずは\figref{figure:exp-target-1}を参考に図の見方を説明する.
  上図は10, 70, 130, 200イテレーション後のサンプリング結果を示しており, 各点はそれぞれBBOまたはLLMによるサンプリング結果を示す.
  x軸は$E^{pos}$, y軸は$E^{torque}$を示しており, パレート解は別の色で表示されている.
  200イテレーション後の解から二つのパレート解Solution 1とSolution 2を選び, それぞれの設計と各指令動作点に逆運動学を解いた際の結果を下図に示している.

  Target-1の実験結果について述べる.
  各イテレーションでパレート解の多くをLLMによるサンプリングが占めていることがわかる.
  Solution 1は$E^{pos}$が小さく$E^{torque}$が大きい解であり, Solution 2はその逆である.
  Solution 1は$\bm{O}$を一番全ての動作指令点にアクセスしやすい原点をとし, 関節配置を$Y-P-R-P$ ($R$をRoll, $P$をPitch, $Y$をYawとした)として全動作点にまんべんなく到達出来ている.
  一方でSolution 2は$\bm{O}$を$z$軸の負方向にずらし, 関節配置を$Y-Y-P-Y$として, 指令動作点への到達度は低いがなるべくトルクがかからないような構造になっている.

  ここで, 例としてSolution 1がサンプリングされたときのLLMの出力の一部を\figref{figure:exp-target-1-llm}に示す.
  Step 1で各パラメータの特性を, Step 2で現在のパレート解の問題を分析し, Step 3で実際にパラメータを出力している.
  Step 2でのJoint Arrangementに関する分析では, それまで複数の同じ関節タイプを利用する解が多かったが, それを明示的に変更して$Y-P-R-P$という解を出力している.
  また, Step 3のORIGINの出力では, 対称なワークスペースを確保するために$\bm{O}$をぴったり原点に設定している.
  特にこの対称なワークスペースの確保は重要であり, このTarget-1という問題設定は$\bm{O}$を原点に設定することで非常に問題が解きやすくなる.
  このような推論はブラックボックス最適化では難しいが, LLMは問題設定から適切な$\bm{O}$の位置を設定することができ, その有効性が確認できる.
}%

\subsection{Target-2 Experiment} \label{subsec:exp-2}
\switchlanguage%
{%
  We describe the experimental results for Target-2 (\figref{figure:exp-target-2}).
  In the case of Target-2, LLM-based sampling also accounts for many of the Pareto-optimal solutions, confirming its effectiveness.
  Unlike Target-1, the joint structure for both Solution 1 and Solution 2 is the same, $Y-R-P-R$.
  However, the second and fourth links in Solution 1 are longer compared to those in Solution 2.
  Due to this difference, Solution 1 achieves a lower $E^{pos}$, while Solution 2 achieves a lower $E^{torque}$.
  In Solution 2, the lengths of the first and second links are set to approximately 0.5 m, allowing the second link to match the height of the target operation points.
  The remaining joints are then used in a manner similar to a Selective Compliance Assembly Robot Arm-type (SCARA-type) structure, helping to minimize the required torque.
}%
{%
  Target-2の実験結果について述べる.
  ここでもLLMによるサンプリングはパレート解の多くを占めており, その有効性が確認できる.
  今回はTarget-1のときとは異なり, Solution 1とSolution 2で関節構造は$Y-R-P-R$と同じである.
  しかし, Solution 1の方がSolution 2に比べて2つめと4つめのリンクが長い.
  この違いによって, Solution 1は$E^{pos}$が小さく, Solution 2は$E^{torque}$が小さいという結果になっている.
  Solution 2は1つめのリンクと2つめのリンクの長さを約0.5 mに設定することで, 2リンク目までで指令動作点への高さを合わせ, 残りの関節をスカラ型のように使うことでトルクを抑える構造になっている.
}%

\subsection{Target-3 Experiment} \label{subsec:exp-3}
\switchlanguage%
{%
  We describe the experimental results for Target-3 (\figref{figure:exp-target-3}).
  For Target-3, while LLM-based sampling contributes to the results, the effect of BBO-based sampling is more prominent.
  Solution 1 adopts a joint arrangement of $Y-P-P-Y$ and places $\bm{O}$ near the center of all target operation points, thereby minimizing $E^{pos}$.
  In contrast, Solution 2 uses a joint arrangement of $Y-Y-P-Y$, reducing the number of pitch axes that are prone to high torque, and shifts $\bm{O}$ further in the negative $z$-axis direction to minimize $E^{torque}$.
  The approach to constructing the solutions is similar to that of Target-1.
}%
{%
  Target-3の実験結果について述べる.
  ここではLLMも寄与しているが, BBOによるサンプリングの効果が大きい.
  Solution 1は関節配置を$Y-P-P-Y$とし, $O$を全ての指令動作点の中心に近い位置に設定することで$E^{pos}$を小さくしている.
  一方でSolution 2は関節配置を$Y-Y-P-Y$としてトルクの受けやすいピッチ軸を減らし, $\bm{O}$をよりz軸の負方向へ持ってくることで$E^{torque}$を小さくしている.
  解の構築方法自体はTarget-1と似通っている.
}%

\section{Discussion} \label{sec:discussion}
\switchlanguage%
{%
  We discuss the experimental results obtained in this study.
  First, we summarize the overall trends.
  LLM-based sampling was effective across multiple problems, although the degree of effectiveness varied for each.
  Additionally, the performance was found to be influenced by the ratio of LLM-based sampling to BBO-based sampling.
  Furthermore, the performance of LLM-based sampling varied significantly depending on the prompt configuration.
  In particular, by incorporating the specific parameter analysis instructions proposed in this study, better design solutions could be explored.
  It was also observed that LLM-based sampling enabled solutions that are difficult for black-box optimization alone, such as placing $\bm{O}$ precisely at the origin or ensuring a symmetric workspace along the $xz$-plane.
  Additionally, it was confirmed that LLM-based sampling reduced randomness and produced more stable results.
  However, it was also noted that values such as link lengths tended to be output as rounded values, making it difficult to perform fine-tuned adjustments.

  Next, we consider the execution time of LLM-based sampling.
  Although LLM inference has gradually become faster, each inference still takes approximately 1-2 seconds.
  However, this issue becomes negligible if LLM-based sampling is executed fully in parallel with BBO.
  In many cases, performance improvements were observed when LLM-based sampling was performed even once for every nine BBO sampling steps.
  This suggests that running LLM-based sampling in parallel, without much consideration, could potentially lead to more efficient sampling.

  Finally, we discuss the limitation of the proposed method.
  In this study, we adopted a relatively straightforward and general problem setting.
  However, in real-world robot design, it is often necessary to consider more complex constraints, joint structures, and link geometries.
  The extent to which LLMs can contribute to design optimization depends on their spatial reasoning capabilities and the complexity of the problem setting.
  There remain several challenges with LLMs, including hallucinations, insufficient spatial reasoning capability, and high sensitivity to prompt variations.
  As these issues are addressed in future developments, the applicability of LLMs to design optimization is expected to further expand.
  While this study primarily focused on demonstrating a new concept, future work will aim to develop a system that can be applied to more complex problem settings.
  In addition, we believe that similar approaches can be extended to Neural Architecture Search (NAS) \cite{zoph2017nas} and to the joint optimization of design and control, which we identify as promising directions for future research.
}%
{%
  本研究で得られた実験結果について考察する.
  まずは全体的な傾向についてまとめる.
  LLMによるサンプリングは複数の問題において効果的であったが, その効果の度合いはそれぞれ異なる.
  また, LLMによるサンプリングとBBOによるサンプリングの割合もその性能に影響を与えることがわかった.
  加えてLLMによるサンプリングはプロンプトによってその性能が大きく変わり, 特に本研究で提案した具体的なパラメータ分析指示を加えることで, より良い設計解を探索することができる.
  LLMによるサンプリングは, ブラックボックス最適化では難しいような, 例えば$O$をピッタリ原点に置いたり, xz平面に対称なワークスペースを確保できたりすることがわかった.
  また, ランダムネスを減少させ, 安定した結果を得ることができることも確認できた.
  一方でリンク長さなどはキリの良い値しか出力されないため, 最終的な調整まで行うことは難しい.

  次にLLMによるサンプリングの実行時間について考察する.
  LLMの推論速度は速くなっているとは言えど, 一度の推論に1-2秒程度かかってしまう.
  一方で, この問題はBBOと完全に並列にLLMによるサンプリングを実行すれば, 時間自体は気にならなくなる.
  今回ブラックボックス最適化9回に対して1回でもLLMによるサンプリングを実行すれば性能の向上に繋がるケースが多く見られた.
  深いことを考えずにLLMによるサンプリングを並列実行することは, 効率の良いサンプリングに繋がる可能性が高い.

  最後に, 本手法の限界について考察する.
  本手法ではかなり一般的で比較的分かりやすい問題設定を採用した.
  一方で, 実際にロボットを設計しようと考えると, より複雑な制約や関節構造, リンク形状などを考慮する必要があることも多い.
  LLMの空間把握能力と問題設定の難しさによってどの程度LLMが設計最適化というドメインに寄与できるかが決まる.
  まだLLMのハルシネーションや空間推論能力の不足, プロンプト変化に対する感度の高さなどの課題もあるため, 今後それらの改善に伴い, さらにLLMの設計最適化への応用範囲は広がっていくだろう.
  本研究は新しいコンセプトを示すことに終始したが, 今後はより複雑な問題設定に対しても適用できるようなシステムの構築を目指す予定である.
  また, 同様の手法がNeural Architecture Search (NAS) \cite{zoph2017nas}や, 制御との共同最適化などにも応用できると考えられ, 今後の研究課題としたい.
}%

\section{CONCLUSION} \label{sec:conclusion}
\switchlanguage%
{%
  In this study, we proposed a robot design optimization method that leverages LLM-based sampling alongside black-box optimization.
  By providing broad design guidelines through LLMs, it is possible to output solutions more efficiently than before, thereby improving the sampling efficiency of black-box optimization.
  LLM-based sampling needs to be used in an appropriate ratio alongside black-box optimization, and the configuration of the LLM prompt has a significant impact on the results.
  In particular, adding specific parameter analysis instructions enables the exploration of better design solutions.
  This approach makes it possible to explore symmetric design solutions, which are challenging for black-box optimization alone, and it has been shown to be highly effective depending on the problem setting.
  In the future, we aim to develop a system that can quickly propose personalized robot body designs based on user tasks and preferences.
}%
{%
  本研究では, LLMによるサンプリングとブラックボックス最適化を活用したロボット設計最適化手法を提案した.
  ブラックボックス最適化のサンプリング効率を上げるために, 大まかな設計指針をLLMにより与えることで, これまでよりも効率的に解を出力することが可能である.
  LLMのサンプリングはブラックボックス最適化に対して適切な割合で併用する必要があり, そのプロンプトの設定は大きな影響を与える.
  特に具体的なパラメータ分析指示を加えることで, より良い設計解を探索することができる.
  ブラックボックス最適化では難しい対称な設計解の探索が可能となり, 問題設定によっては絶大な効果を発揮することがわかった.
  今後, ユーザにパーソナライズされたロボット身体設計をタスクや好みから素早く提案できるようなシステムの構築を目指す.
}%

\section*{Acknowledgment}
We utilized GPT-5 to assist in revising the manuscript.

\bibliographystyle{IEEEtran}
\bibliography{main}

\begin{IEEEbiography}[{\includegraphics[width=1in,height=1.25in,clip,keepaspectratio]{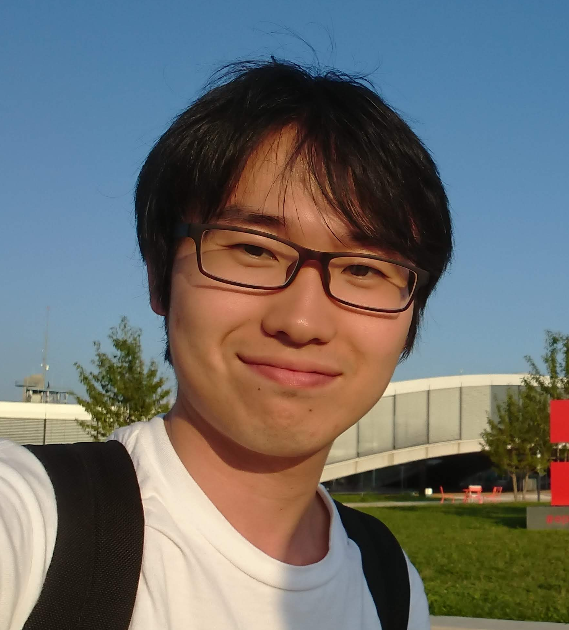}}]{K. Kawaharazuka} received his B.E. in Mechano-Informatics from The University of Tokyo in 2017. He received his M.S. in Information Science and Technology from The University of Tokyo in 2019, and his Ph.D. in the same field in 2022. In 2022, he was appointed as a Project Assistant Professor at the Graduate School of Information Science and Technology, The University of Tokyo. Since 2025, he has been serving as a Lecturer at the Next Generation Artificial Intelligence Research Center, Graduate School of Information Science and Technology, The University of Tokyo. His research interests include the body design and control of musculoskeletal humanoids, as well as intelligent robotic systems based on deep learning and foundation models.
\end{IEEEbiography}

\begin{IEEEbiography}[{\includegraphics[width=1in,height=1.25in,clip,keepaspectratio]{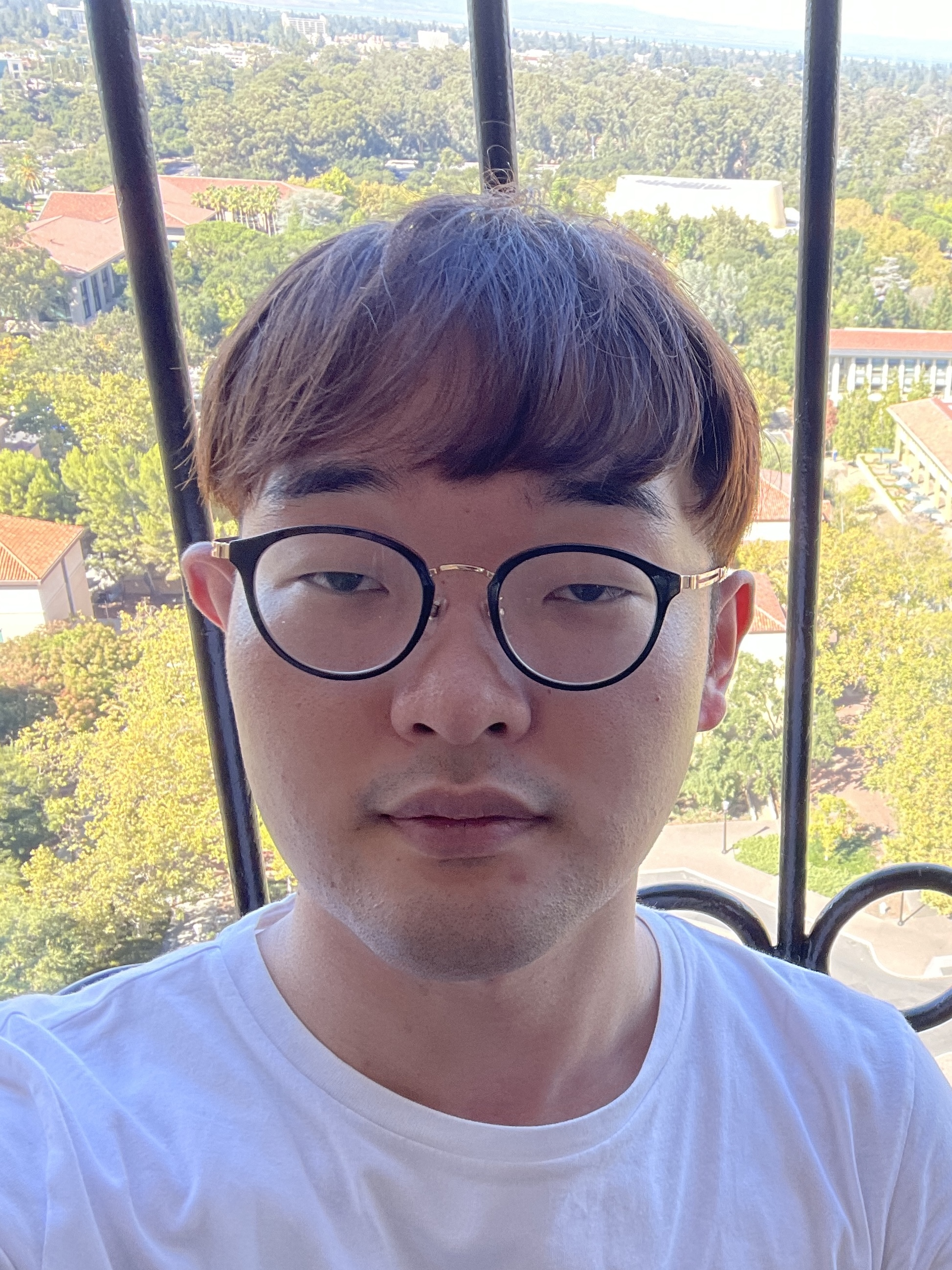}}]{Y. Obinata} received his B.E. in the Department of Mechano-Informatics from The University of Tokyo in 2021. He received his M.S. in Information Science and Technology from The University of Tokyo in 2023. He is a Ph.D. course student at the Graduate School of Information Science and Technology, The University of Tokyo. He was awarded the IEEE Robotics and Automation Society Japan Joint Chapter Young Award and the SICE International Young Authors Award in 2023. His research interests include robot system integration and developing the robot communication system.
\end{IEEEbiography}

\begin{IEEEbiography}[{\includegraphics[width=1in,height=1.25in,clip,keepaspectratio]{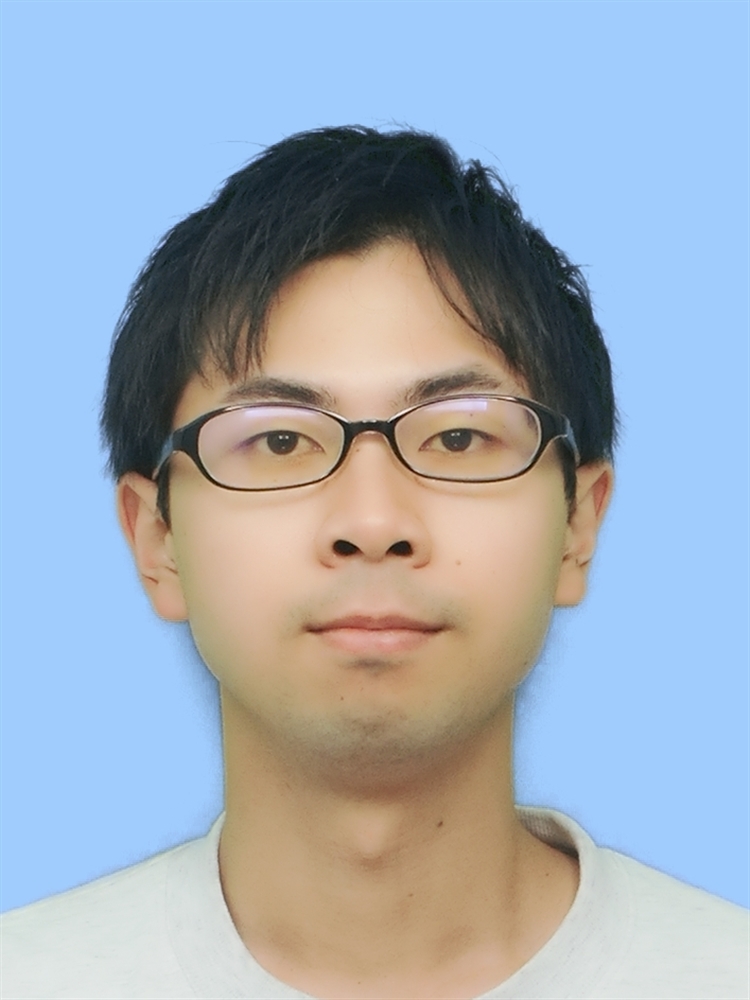}}]{N. Kanazawa} is a Ph.D. course student at JSK Robotics Laboratory in the Department of Mechano-Informatics at the University of Tokyo. He received his B.E. and M.S. degrees in Mechano-Informatics from the University of Tokyo in 2021 and 2023, respectively. His research interests include cooking robot systems.
\end{IEEEbiography}

\begin{IEEEbiography}[{\includegraphics[width=1in,height=1.25in,clip,keepaspectratio]{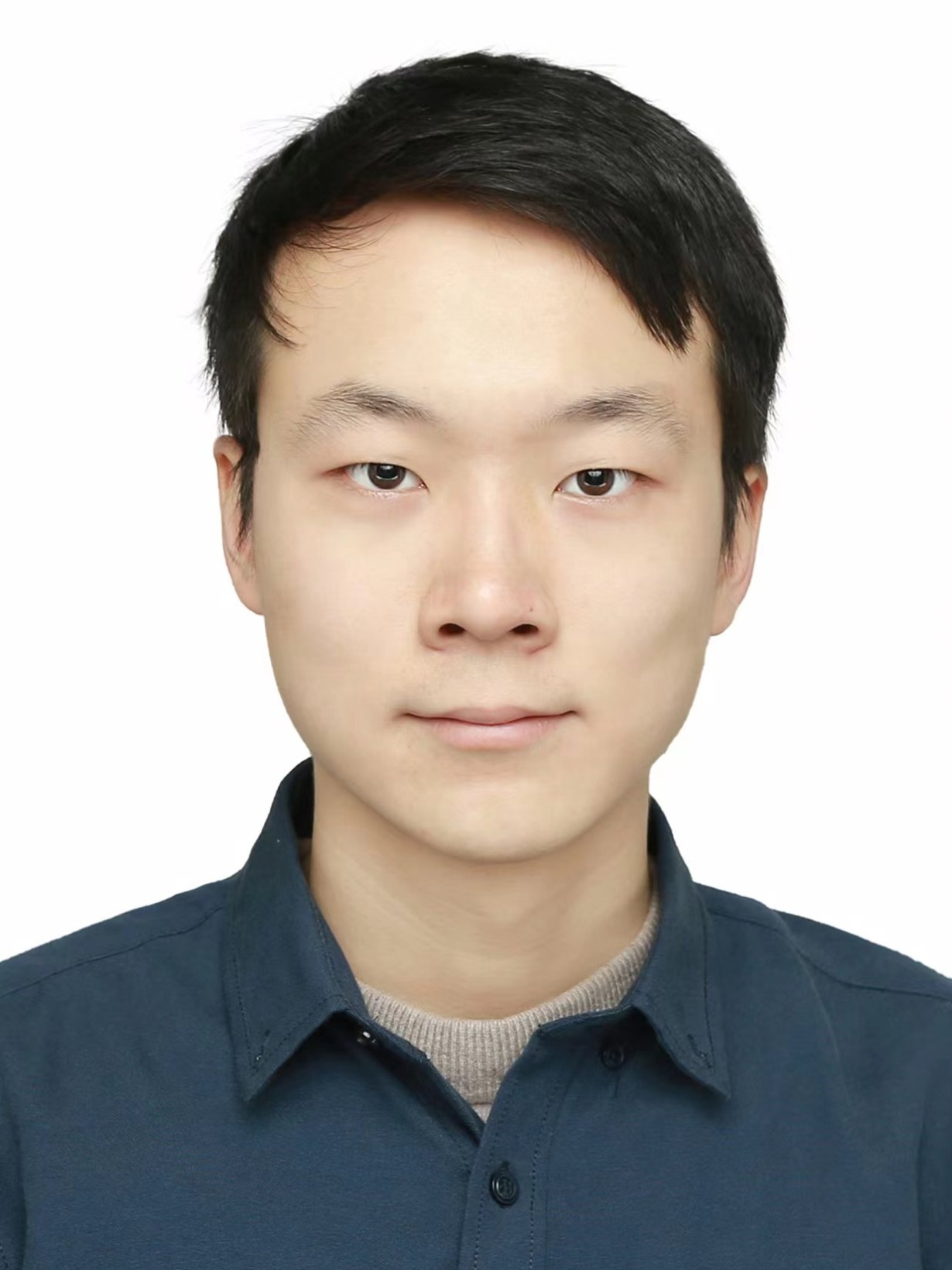}}]{H. Jia} is a M.S course student, affiliated with JSK Robotics Laboratory in the Department of Mechano-Informatics at the University of Tokyo. His B.S. degree of Data Science is granted by the Science Department of China University of Petroleum. He is dedicated to research on LLM-driven intelligent systems for robot agents.
\end{IEEEbiography}

\begin{IEEEbiography}[{\includegraphics[width=1in,height=1.25in,clip,keepaspectratio]{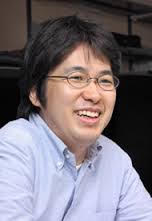}}]{K. Okada} received his B.E. in Computer Science from Kyoto University in 1997. He received his M.S. and Ph.D. in Information Engineering from The University of Tokyo in 1999 and 2002, respectively. From 2002 to 2006, he joined the Professional Programme for Strategic Software Project in The University Tokyo. He was appointed as a lecturer in the Creative Informatics at the University of Tokyo in 2006, an associate professor and a professor in the Department of Mechano-Informatics in 2009 and 2018, respectively. His research interests include humanoids robots, real-time 3D computer vision, and recognition-action integrated system.
\end{IEEEbiography}

\EOD

\end{document}